\documentclass[runningheads]{llncs}

\usepackage{eccv}

\usepackage{eccvabbrv}

\usepackage{graphicx}
\usepackage{booktabs}
\usepackage[dvipsnames]{xcolor}
\usepackage[table]{xcolor}
\usepackage{pifont}
\usepackage{multirow}
\newcommand{\cmark}{\ding{51}}
\newcommand{\xmark}{\ding{55}}
\usepackage{tcolorbox}
\usepackage{tcolorbox}
\tcbuselibrary{listings, breakable, skins}
\usepackage{listings}

\usepackage{varwidth}
\usepackage{xcolor}
\definecolor{promptgray}{gray}{0.96}
\definecolor{keycolor}{RGB}{0, 70, 127}
\usepackage{tcolorbox}
\tcbuselibrary{listings, breakable, skins}
\usepackage{xcolor}

\definecolor{promptgray}{gray}{0.96}
\definecolor{keycolor}{RGB}{0, 70, 127}

\tcbset{
  promptbox/.style={
    breakable,
    colback=gray!5,
    colframe=gray!40,
    fonttitle=\small\bfseries,
    left=4pt, right=4pt, top=3pt, bottom=3pt,
    boxrule=0.4pt,
    arc=2pt,
    listing only,
    listing options={
      basicstyle=\ttfamily\small,
      breaklines=true,
      keepspaces=true,
      columns=flexible,
    },
  }
}

\usepackage[accsupp]{axessibility}  
\usepackage{enumitem}

\usepackage[pagebackref,breaklinks,colorlinks,citecolor=eccvblue]{hyperref}
\usepackage{hyperref}

\usepackage{orcidlink}

\begin{document}

\title{EchoSonar-R: A Multi-View Reasoning-Enabled Model for Disease Classification and Report Generation in Echocardiography} 

\titlerunning{EchoSonar-R}

\author{Darya Taratynova\inst{1}\orcidlink{0009-0005-8344-7709} \and
Ahmed Aly\inst{1}\orcidlink{0009-0000-1133-2463} \and
Numan Saeed\inst{1}\orcidlink{0000-0002-6326-6434}$^{\star}$ \and
Mohammad Yaqub\inst{1}\orcidlink{0000-0001-6896-1105}$^{\star}$}

\authorrunning{Taratynova et al.}

\institute{Division of Computing and Mathematical Sciences, Mohamed Bin Zayed University of Artificial Intelligence (MBZUAI), Abu Dhabi, UAE\\
\email{\{darya.taratynova, ahmed.aly, numan.saeed, mohammad.yaqub\}@mbzuai.ac.ae}}

\maketitle
\footnotetext[1]{$^{\star}$ Joint supervision.}

\begin{abstract}
Echocardiography is the most widely used non-invasive cardiac imaging modality, providing essential information for cardiovascular diagnosis. Interpreting an echocardiogram requires synthesizing complementary evidence across multiple heart views to identify abnormalities and produce structured clinical reports. While recent efforts focus on improving classification performance, most models lack explicit diagnostic reasoning and spatially grounded anatomical evidence, limiting clinician trust. We present EchoSonar-R, a multi-view reasoning-enabled vision–language model that jointly performs multi-label disease classification and report generation from echocardiography studies. EchoSonar-R combines a spatiotemporal video encoder with a structure-aware cardiac detector that provides spatially grounded anatomical cues to improve interpretability and clinician trust during cross-view reasoning. EchoSonar-R is trained in two stages: supervised fine-tuning (SFT) on reasoning-annotated targets, followed by Group Relative Policy Optimization (GRPO) with task-specific rewards that jointly align classification and report generation within a unified reinforcement-learning framework. Across a private multi-view dataset and two public benchmarks, EchoSonar-R improves macro balanced accuracy by 17.1\% on the 
private set and 6.1\% on MIMICEchoQA over the strongest baseline, achieves a GREEN clinical faithfulness score of 0.800, and produces interpretable reasoning traces grounded in multi-view visual evidence.

\keywords{Echocardiography \and Multi-view Reasoning \and Reinforcement Learning \and Vision Language Models (VLMs)}
\end{abstract}

\section{Introduction}
\label{sec:intro}

Cardiovascular diseases (CVDs) are the leading cause of death worldwide, 
accounting for 19.8 million deaths in 2022, with over three-quarters of these deaths occurring in low- and middle-income countries (LMICs) ~\cite{who2024cvd,roth2024heartofworld}. The global CVD burden is rising due to population growth, aging, and increasing risk factors, with projections of 35.6 million cardiovascular deaths annually by 2050~\cite{lim2024cvdprojections}. In routine clinical care, echocardiography is the most widely used non-invasive cardiac imaging modality ~\cite{statpearls2023echoimaging,chambers2012echofrontier}. During a comprehensive echocardiographic study, cardiologists acquire and interpret multiple views synthesizing complementary anatomical and functional information to evaluate heart function, documenting their findings in structured clinical reports~\cite{statpearls2023echoimaging,vukadinovic2024echoprime}. Globally, echocardiography demand continues to grow, yet the supply of qualified sonographers is not keeping pace~\cite{won2024sonographershortage}. This disparity is particularly acute in LMICs, where access to echocardiography remains limited ~\cite{escatlas2024cvd,arega2023imagingafrica,savarese2023cvdhealthcare}.
The growing gap between rising diagnostic demand and the global shortage of trained sonographers underscores the need for automated solutions capable of not only performing echocardiographic measurements but also synthesizing multi-view findings into explicit diagnostic reasoning and structured clinical reports, thereby improving the efficiency, consistency, and accessibility of expert-level cardiac assessment worldwide \cite{mitchell2019guidelines}.

\begin{figure}[t]
    \centering
    \includegraphics[width=0.95\linewidth]{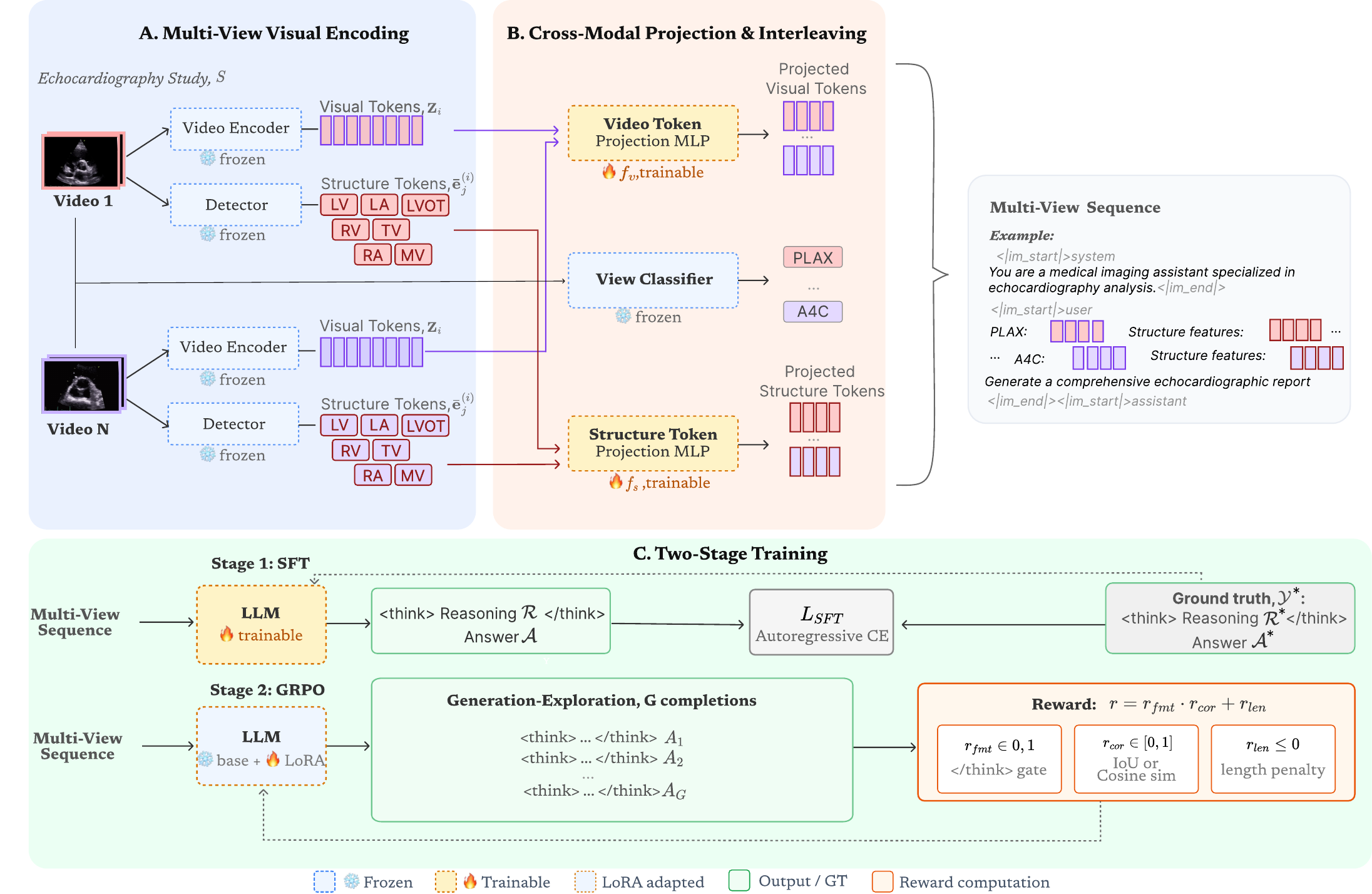}
    \caption{Overview of EchoSonar-R. \textit{Multi-View Visual Encoding:} each video is processed by a frozen spatiotemporal encoder and a frozen structure-aware detector to extract complementary global and anatomical tokens for seven cardiac structures. \textit{Cross-Modal Projection and Interleaving:} visual and structure tokens are projected into the language model embedding space via trainable MLPs and interleaved with view identifiers to form a multi-view input sequence. \textit{Two-Stage Training:} SFT fine-tunes the model on reasoning-annotated targets; GRPO subsequently optimizes a composite reward.}
    \label{fig:architecture}
\end{figure}

Automated echocardiographic analysis has gained momentum to reduce clinician workload and improve diagnostic consistency~\cite{ghorbani2020deeplearningecho,zhang2018fullyautomatedecho}. Early deep learning approaches targeted single-task objectives such as view classification~\cite{jansen2024automated}, chamber segmentation~\cite{maani2024simlvseg}, and ejection fraction estimation~\cite{asch2019automated,li2023echoefnet,rahman2023deep,maani2024coreecho,muhtaseb2022echocotr}. While effective for isolated tasks, these models do not synthesize complementary information from the multiple views, thereby limiting both their diagnostic scope and clinical applicability. More recently, echocardiography foundation models such as EchoCLIP~\cite{christensen2024echoclip}, EchoPrime~\cite{vukadinovic2024echoprime}, and PanEcho~\cite{holste2025panecho} have improved echocardiography representations via large-scale pretraining. However, these models primarily map visual inputs to classification labels or retrieve reports without generating explicit, cross-view grounded reasoning or producing clinician-style reports.  This lack of interpretable reasoning limits trust and hinders adoption in clinical workflows~\cite{salih2023xaicardiacimaging}. This motivates a framework that fuses multi-view echocardiographic evidence, enables diagnostic reasoning, and generates comprehensive clinical reports. In this work, we introduce \textbf{EchoSonar-R}\footnote{\url{https://github.com/BioMedIA-MBZUAI/EchoSonar-R}}, the first multi-view reasoning-enabled model for echocardiographic disease classification and report generation (\cref{fig:architecture}). Our contributions are: 
\begin{itemize}
    \item a multi-view architecture that combines a view-aware video encoder with a structure-aware cardiac detector, enabling cross-view reasoning and providing spatially grounded anatomical cues alongside global motion dynamics for improved clinical trust.
    \item a two-stage training pipeline: supervised fine-tuning (SFT) with reasoning-annotated targets followed by group relative policy optimization (GRPO) with task-specific reward functions, that unifies multi-label classification and report generation within a single reinforcement learning (RL) framework.
    \item extensive experiments on three datasets, demonstrating strong performance on both abnormality classification and report generation.
\end{itemize}
\section{Related Work}
\label{sec:related_work}

\textbf{Automated Echocardiography Report Generation.} Automated report generation has progressed from early CNN-RNN image captioning frameworks~\cite{vinyals2015showandtell,xu2015showattendtell} to transformer-based architectures~\cite{vaswani2017attention}. In radiology, advances span memory-driven transformers~\cite{chen2020r2gen,chen2022r2gengpt}, contrastive anomaly-focused generation~\cite{yan2023contrastive}, and knowledge graph grounding~\cite{li2023knowledgegraph}. In echocardiography specifically, rather than generating text directly from images, existing approaches produce reports from structured measurement inputs provided by sonographers. Syryca et al.~\cite{syryca2025echoreportllm} evaluated ChatGPT's ability to generate comprehensive echocardiography reports from numerical measurements, while Chao et al.~\cite{chao2025echogpt} developed EchoGPT for echo report summarization. However, these methods depend on sonographer-derived measurements and cannot produce integrated reports directly from echocardiographic video.

\noindent \textbf{Medical Vision-Language Models.} Contrastive vision–language pretraining~\cite{radford2021clip} has driven rapid progress in medical VLMs. MedGemma~\cite{sellergren2025medgemma} provides open medical foundation models with a medically tuned SigLIP encoder pretrained across diverse imaging modalities, while Lingshu~\cite{xu2025lingshu} introduces a generalist medical MLLM spanning 12+ modalities and strong medical VQA performance. In ultrasound, EchoVLM~\cite{she2025echovlm} proposes a Mixture-of-Experts model trained across seven anatomical regions for report generation, diagnosis, and VQA. In echocardiography, EchoCLIP~\cite{christensen2024echoclip} aligns echo videos with clinical text via contrastive pretraining for zero-shot assessment, and its successor EchoPrime~\cite{vukadinovic2024echoprime} extends this with a multi-view, study-level architecture for more comprehensive interpretation.


\noindent \textbf{Reasoning-Enabled Medical Models.}
Recent work has begun to incorporate explicit reasoning into medical VLMs, motivated by chain-of-thought (CoT) progress in general-purpose models~\cite{wei2022cot}. MedVLM R1~\cite{pan2025medvlmr1} uses RL to encourage human-interpretable reasoning in medical image analysis, while Chiron-o1~\cite{sun2025chirono1} generates high-quality medical CoT data via curriculum learning, achieving strong results on medical VQA and reasoning benchmarks. However, these approaches largely focus on single-image or single-modality VQA and do not address the multi-view reasoning required in echocardiography. EchoSonar-R addresses this gap by enabling multi-view, reasoning-based report generation and disease classification, grounded in anatomical evidence.

\section{Method}
\label{sec:method}

Given an echocardiographic study $\mathcal{S} = \{V_1, V_2, \ldots, V_N\}$ comprising $N$ videos from distinct standard views and a task prompt $q$, EchoSonar-R generates a structured response 
$\mathcal{Y} = [\langle\texttt{think}\rangle\, \mathcal{R}\, \langle\texttt{/think}\rangle\, \mathcal{A}]$, where $\mathcal{R}$ is a reasoning trace and $\mathcal{A}$ is the answer to $q$.

As illustrated in Fig.~\ref{fig:architecture}, the model consists of three components: (1)~a \textbf{multi-view visual encoder}, processing each video $V_i$ through a frozen spatiotemporal backbone and a structure-aware detector to extract complementary global and anatomical representations, (2)~\textbf{cross-modal projection layers}, mapping the visual tokens into the language model embedding space and interleave them with view identifiers and the task prompt, and (3)~a \textbf{reasoning-enabled language model}, synthesizing 
information across all views to produce $\mathcal{Y}$. 

\subsection{Multi-View Visual Encoding}
\label{sec:encoding}
\textbf{Spatiotemporal Video Encoder.}
Each video $V_i \in \mathcal{S}$ is processed by a frozen mViT-based encoder~\cite{vukadinovic2024echoprime} to obtain a sequence of $T = 393$ visual tokens $\mathbf{Z}_i \in \mathbb{R}^{T \times d_v}$. 
The encoder remains frozen throughout training to preserve its pretrained motion and view-specific priors, leaving cross-modal alignment to projectors. Unlike prior work that operates on single frames or individual views, this formulation retains the full token sequence from each video, allowing the downstream language model to attend to cues across the cardiac cycle.

\noindent\textbf{Structure-Aware Detector.}
While the spatiotemporal encoder captures global cardiac dynamics, echocardiographic interpretation further requires a localized understanding of individual anatomical structures. To this end, we train an RT-DETR-L detector~\cite{zhao2024rtdetr} to localize seven key cardiac structures. For each video $V_i$, we run the detector across frames and extract the learned object query embeddings $\mathbf{e}^{(i)}_j \in \mathbb{R}^{d_s}$, corresponding to each detected structure $j$. Query embeddings for the same structure are averaged across frames within a video to obtain a single per-structure representation $\bar{\mathbf{e}}^{(i)}_j \in \mathbb{R}^{d_s}$, capturing the characteristic appearance of each anatomical region. This results in up to $K$ structure tokens per video, providing the language model with spatially grounded anatomical cues, and allowing its outputs to be accompanied by corresponding bounding-box evidence.

\noindent\textbf{Cross-Modal Projection and Multi-View Interleaving.}
The visual representations from the spatiotemporal encoder and the structure-aware detector reside in different feature spaces and must be aligned with the language model's embedding space. We introduce two learned projection modules: a video projection $f_v: \mathbb{R}^{d_v} \rightarrow \mathbb{R}^{d_\text{LM}}$ that maps 
each video token, and a structure projection $f_s: \mathbb{R}^{d_s} \rightarrow \mathbb{R}^{d_\text{LM}}$ that maps each detector query embedding, to $d_\text{LM}$ hidden dimension of the language model. Both $f_v$ and $f_s$ are two-layer MLPs with GELU activation. We construct the input sequence by interleaving projected visual tokens with textual view identifiers, which are obtained by a pretrained view classifier. Concretely, for each video $V_i$ acquired from view $v_i$, we prepend a text token sequence encoding the view name, followed by the $T$ projected video tokens and the projected structure tokens. 

\subsection{Stage 1: SFT}
\label{sec:sft}
During SFT, the model learns to interpret the projected multi-view visual tokens, align them with clinical concepts, and generate structured text. This initialization ensures that RL begins with an informative policy operating over a well-calibrated visual input space, thereby stabilizing reward-based optimization~\cite{ouyang2022training, shao2024deepseekmath}.

\noindent\textbf{Reasoning Data Construction.}
For each training study $\mathcal{S}$, we construct ground-truth target sequences $\mathcal{Y}^* = [\langle\texttt{think}\rangle\, \mathcal{R}^*\, \langle\texttt{/think}\rangle\, \mathcal{A}^*]$, pairing a reasoning trace $\mathcal{R}^*$ with a reference answer $\mathcal{A}^*$. Each $\mathcal{A}^*$ is derived from the ground-truth expert report, and $\mathcal{R}^*$ is generated by prompting Qwen3-8B~\cite{yang2025qwen3} with guideline-based templates~\cite{mitchell2019guidelines} to produce a step-by-step diagnostic justification that logically leads to $\mathcal{A}^*$. To encourage diverse reasoning and prevent overfitting to a single output format, we construct training pairs across five complementary question types: (i)~full report generation, (ii)~conclusion-only generation, (iii)~description of a specific cardiac structure, (iv)~binary classification of a specific abnormality, and (v)~multi-label classification. This multi-task formulation encourages the model to develop shared reasoning capabilities that generalize across clinical queries. Full details on the prompt templates and example question-answer pairs are provided in the Supplementary Material.

\noindent\textbf{Training Objective.}
We optimize the autoregressive cross-entropy loss over the target tokens, conditioned on the multi-view visual input and the task prompt:
\begin{equation}
  \mathcal{L}_{\text{SFT}} = 
    -\sum_{t=1}^{|\mathcal{Y}|} 
    \log\, p_\theta\!\left(y_t \mid y_{<t},\, \mathbf{X},\, q\right),
\end{equation}
where $\mathbf{X}$ is the interleaved multi-view input sequence and $\theta$ denotes all trainable parameters. During SFT, we unfreeze both the cross-modal projection layers ($f_v$, $f_s$) and the full language model parameters, while the visual encoders remain frozen. 

\subsection{Stage 2: Group Relative Policy Optimization (GRPO)}
\label{sec:grpo}

We apply GRPO~\cite{shao2024deepseekmath} to move beyond token-level imitation of generated reasoning, which does not directly optimize for task-level correctness~\cite{xu2026reasoning}. GRPO instead rewards completions based on their final clinical correctness, allowing the model to refine its decision boundaries through outcome-based exploration.

For each training prompt $x$, we sample $G$ independent completions 
$\{y_1, \ldots, y_G\}$. Each completion receives a scalar reward $r_i$, from which we derive group-normalized advantages $\hat{A}_i = \frac{r_i - \bar{r}}{\sigma_r + \epsilon}$, where $\bar{r}$ and $\sigma_r$ are the mean and standard deviation of rewards within the group. The policy is updated by maximizing the advantage-weighted log-likelihood over completion tokens:
\begin{equation}
    \mathcal{L}_{\text{GRPO}} = 
    -\frac{1}{G} \sum_{i=1}^{G} \hat{A}_i \cdot 
    \frac{1}{T_i} \sum_{t=1}^{T_i} 
    \log \pi_\theta(y_t \mid x, y_{<t}).
\end{equation}
Following recent work on reasoning models~\cite{yu2025dapo, liu2025understanding}, we omit the KL divergence penalty from the standard GRPO objective. The KL term regularizes the policy toward the reference model, potentially constraining exploration~\cite{yu2025dapo}.

\textbf{Reward design.}
The composite reward balances structural validity, correctness, and output completeness:
\begin{equation}
    r = r_{\text{fmt}} \cdot r_{\text{cor}} + r_{\text{len}}.
\end{equation}

\noindent\textbf{Format reward}, $r_{\text{fmt}} \in \{0, 1\}$. The reward equals 1 if the completion contains \texttt{</think>} tag, 
and 0 otherwise. This gates the entire correctness signal, enforcing 
well-formed chain-of-thought structure as a prerequisite for receiving any credit. 

\noindent\textbf{Correctness reward}, $r_{\text{cor}} \in [0, 1]$. The correctness reward $r_{\text{cor}} \in [0, 1]$ is task-dependent and designed to provide a smooth, 
differentiable training signal. 

\noindent\emph{(i) Multi-label classification.} We compute:
\begin{equation}
    r_{\text{cor}}^{\text{multi-label}} = 
    \frac{|\mathcal{P} \cap \mathcal{G}|}{|\mathcal{P} \cup \mathcal{G}|},
\end{equation}
where $\mathcal{P}$ and $\mathcal{G}$ denote the predicted and ground-truth sets of abnormalities.  IoU provides a graded signal that rewards partial correctness and penalizes both missed findings (false negatives in $\mathcal{G} \setminus \mathcal{P}$) and hallucinated findings (false positives in $\mathcal{P} \setminus \mathcal{G}$) symmetrically.

\noindent\emph{(ii) Report generation.} We compute per-section semantic similarity using the pretrained echocardiography-aware text encoder:
\begin{equation}
    r_{\text{cor}}^{\text{report}} = 
    \frac{1}{|\mathcal{C}|} \sum_{c \in \mathcal{C}} 
    \cos\!\big(\mathbf{e}_c^{\text{pred}},\, 
    \mathbf{e}_c^{\text{gt}}\big)
    \label{eq:report_reward},
\end{equation}
where $\mathcal{C}$ is the set of report sections, and $\mathbf{e}_c^{\text{pred}}$, $\mathbf{e}_c^{\text{gt}}$ are the text encoder embeddings of the predicted and ground-truth text for section $c$. Missing predicted sections are assigned a similarity of zero. We adopt semantic similarity rather than lexical metrics such as BLEU because echocardiographic reports allow substantial linguistic variation for clinically equivalent findings (e.g., “mildly dilated left atrium” vs. “LA is mildly enlarged”), which $n$-gram overlap metrics penalize despite identical clinical meaning. A specialized clinical text encoder better captures these domain-specific semantic equivalences.

\noindent\textbf{Length penalty}, $r_{\text{len}}\le 0$. The length penalty discourages short completions:
$r_{\text{len}}=\min\!\left(0,\frac{L-L_{\min}}{L_{\min}}\right)$,
where $L$ is the completion length (tokens) and $L_{\min}=400$. Completions shorter than $L_{\min}$ receive a penalty proportional to their deficit, while longer ones receive no penalty.
\section{Experiments}
\label{sec:intro}

EchoSonar-R is evaluated on two core tasks, abnormality classification and report generation, across three datasets spanning private multi-view and public single-view echocardiography:

\noindent\textit{\textbf{Private Dataset:}} We train and evaluate on a private echocardiography dataset with 5,061 training and 1,215 test studies, comprising 46,382 and 11,310 videos, respectively. Each study includes a median of 7 multi-view videos, with a median of 31 frames per video. The dataset covers 12 abnormality categories spanning valvular disease, chamber enlargement, systolic dysfunction, valve calcification, and structural anomalies, with prevalence ranging from 4.4\% to 65.0\%, reflecting natural class imbalance. Each report is structured into $\mathcal{C}$=13 section-level descriptions followed by a clinical conclusion. For the structure-aware detector, a subset of 2,689 studies was annotated with segmentation masks for seven cardiac structures, converted to bounding boxes and propagated across frames using RAFT optical flow~\cite{teed2020raft}. Full dataset statistics are provided in the Supplementary Material.
 
\noindent\textit{\textbf{MIMICEchoQA~\cite{thapa2025mimicechoqa}:}} We additionally evaluate on MIMICEchoQA, a publicly available echocardiogram-based visual question-answering benchmark derived from the MIMIC-IV-ECHO database~\cite{gow2023mimicivecho, physionet2000}. The dataset comprises 622 clinician-validated, closed-ended QA pairs, each linked to a unique single-view echocardiogram video. We convert the closed-ended QA pairs into binary abnormality classification labels, retaining the subset of 155 samples pertaining to five of our target conditions. 

\noindent\textit{\textbf{EchoNet-Dynamic~\cite{ouyang2020echonetdynamic}:}} We further evaluate on EchoNet-Dynamic, a large publicly available echocardiography dataset comprising A4C view videos collected during routine clinical care at Stanford University Hospital. Each video is paired with expert-annotated ejection fraction (EF) and left ventricular tracings. We use the standard test split of 1,277 studies and convert continuous EF values into a binary LV systolic dysfunction label: studies with EF in the range $[55\%, 70\%]$ are labeled as normal, following established clinical guidelines that define normal LV systolic function within this range~\cite{lang2015ase}.

\noindent\textbf{Evaluation Metrics.} We evaluate our model on two tasks: (i)~\emph{abnormality classification}, and (ii)~\emph{report generation}. Following prior work in medical image classification~\cite{realenosei2024medicalcaptionsurvey, vukadinovic2024echoprime}, we report \textbf{per-disease F1 scores}, \textbf{balanced accuracy}, and \textbf{macro-averaged metrics} across all abnormality categories. We adopt balanced accuracy and macro-averaging rather than standard accuracy to account for the substantial class imbalance in our dataset, ensuring that performance on rare conditions contributes equally to the overall evaluation.

To evaluate the quality of the reasoning traces produced within the \texttt{<think>} blocks, we employ an LLM-as-a-judge framework~\cite{zheng2024judging}. Each reasoning trace is scored on a 1--5 Likert scale across five dimensions: ~\textbf{Reasoning--Answer Agreement}, measuring whether every abnormality discussed in the reasoning trace is reflected in the final answer and vice versa; ~\textbf{Reasoning Efficiency}, assessing whether the reasoning is concise and focused, without redundant or circular steps; ~\textbf{Factual Correctness}, measuring the accuracy of clinical and anatomical claims made during reasoning; ~\textbf{Evidence Grounding}, evaluating whether the model references specific visual observations from the echocardiographic input rather than relying on generic statements; and ~\textbf{Terminology Accuracy}, assessing the correct use of echocardiographic and medical terminology throughout the reasoning chain.

The generated reports are evaluated using natural language generation (NLG) metrics: \textbf{BLEU-1/2/3/4}~\cite{papineni2002bleu}, which measure $n$-gram precision between generated and reference reports; \textbf{METEOR}~\cite{banerjee2005meteor}, which extends $n$-gram matching with synonym recognition and stemming; and \textbf{ROUGE-L}~\cite{lin2004rouge}, which captures the longest common subsequence between candidate and reference texts. However, these lexical overlap metrics are known to correlate poorly with clinical correctness~\cite{ostmeier2024green}, as semantically equivalent medical descriptions may use different surface forms. To address this, we compute \textbf{BERTScore}~\cite{zhang2020bertscore}, which evaluates semantic similarity via token-level cosine similarities between contextual embeddings of candidate and reference reports. Furthermore, we report the \textbf{GREEN} score~\cite{ostmeier2024green}, which identifies and classifies clinically significant errors in generated reports. It produces a score between 0 and 1 that has been shown to align more closely with expert preferences than conventional NLG metrics~\cite{ostmeier2024green}.

\noindent \textbf{Implementation Details.}
The spatiotemporal encoder is a frozen EchoPrime 
(mViT)~\cite{vukadinovic2024echoprime}, $d_v = 768$; we also use its 
pretrained text encoder to compute the per-section semantic 
similarity reward in \cref{eq:report_reward}. The detector is RT-DETR-L~\cite{zhao2024rtdetr} trained for 30 epochs with class-balanced sampling,  $d_s = 256$, $K = 7$. Inputs are 16 frames per video with stride of 2. The language model is Qwen3-8B~\cite{yang2025qwen3}. SFT is trained for 3 epochs (LR $2{\times}10^{-5}$, AdamW optimizer, batch size 64). GRPO uses $G{=}8$ completions ($\tau{=}1$, top-$p$ $0.95$), LoRA ($r{=}64$, $\alpha{=}128$) on all LM projections with frozen base weights, LR $1{\times}10^{-6}$, cosine schedule, 5 epochs, DeepSpeed ZeRO-2. BERTScore is computed using PubMedBERT~\cite{gu2021pubmedbert}. Reasoning quality and GREEN score are computed using Mistral-7B~\cite{jiang2023mistral}.

\begin{table*}[t]
\centering
\caption{Abnormality classification performance across private and 
public datasets. We report per-disease F1 score ($\uparrow$) and 
balanced accuracy (BAcc, $\uparrow$). EchoSonar-R$^{\dagger}$ 
denotes GRPO training; EchoSonar-R$^{\ast}$ denotes SFT-only. 
Baselines: Qwen3-VL~\cite{bai2025qwen3vl}, 
MedGemma~\cite{sellergren2025medgemma}, 
EchoVLM~\cite{she2025echovlm}, 
Chiron-o1~\cite{sun2025chirono1}, 
Lingshu~\cite{xu2025lingshu}. 
Best results in \textbf{bold}, second best \underline{underlined}. 
Prev.\ = prevalence (\%).}
\label{tab:classification}
\setlength{\tabcolsep}{4.5pt}
\resizebox{\textwidth}{!}{%
\begin{tabular}{l c | cc | cc || cc | cc | cc | cc | cc}
\toprule
& & \multicolumn{2}{c|}{\textbf{EchoSonar-R}$^{\dagger}$} & \multicolumn{2}{c||}{\textbf{EchoSonar-R}$^{\ast}$} & \multicolumn{2}{c|}{\textbf{Qwen3-VL}} & \multicolumn{2}{c|}{\textbf{MedGemma}} & \multicolumn{2}{c|}{\textbf{EchoVLM}} & \multicolumn{2}{c|}{\textbf{Chiron-o1}} & \multicolumn{2}{c}{\textbf{Lingshu}} \\
\cmidrule(lr){3-4} \cmidrule(lr){5-6} \cmidrule(lr){7-8} \cmidrule(lr){9-10} \cmidrule(lr){11-12} \cmidrule(lr){13-14} \cmidrule(lr){15-16}
\textbf{Abnormality} & \textbf{Prev.} & F1 & BAcc & F1 & BAcc & F1 & BAcc & F1 & BAcc & F1 & BAcc & F1 & BAcc & F1 & BAcc \\
\midrule
\multicolumn{16}{l}{\cellcolor{yellow!8}\textit{\textbf{Private Test Set} ($n = 1{,}215$ studies, multi-view)}} \\[3pt]
TV Regurgitation              & 54.6 & 66.6 & \textbf{59.9} & 66.6 & \underline{59.3} & 38.3 & 52.1 & 27.9 & 51.2 & 0.0  & 50.0 & 61.1 & 48.8 & 62.3 & 49.4 \\[1.5pt]
MV Regurgitation              & 54.4 & 68.3 & \underline{61.6} & 69.1 & \textbf{62.6} & 49.7 & 51.5 & 24.6 & 49.2 & 0.0  & 50.0 & 69.6 & 50.5 & 33.6 & 50.5 \\[1.5pt]
LA Enlargement                & 27.3 & 64.9 & \textbf{75.8} & 62.7 & \underline{74.6} & 26.6 & 49.8 & 35.1 & 49.3 & 0.0  & 50.0 & 40.2 & 49.5 & 29.5 & 48.0 \\[1.5pt]
Healthy                       & 17.5 & 38.0 & \textbf{62.3} & 36.2 & \underline{61.3} & 20.1 & 49.9 & 23.7 & 48.1 & 0.0  & 50.0 & 0.0  & 49.9 & 1.9  & 50.4 \\[1.5pt]
LV Systolic Dysfunction       & 16.3 & 59.5 & \textbf{74.0} & 55.9 & \underline{71.9} & 21.8 & 51.8 & 19.6 & 48.6 & 0.0  & 50.0 & 27.9 & 50.5 & 28.0 & 51.2 \\[1.5pt]
AV Regurgitation              & 15.4 & 42.3 & \textbf{65.4} & 41.3 & \underline{65.1} & 24.2 & 52.3 & 16.8 & 50.4 & 26.8 & 50.0 & 26.1 & 50.0 & 25.6 & 49.9 \\[1.5pt]
AV Stenosis                   &  9.5 & 71.6 & \textbf{82.2} & 69.8 & \underline{80.9} & 14.9 & 50.3 & 10.0 & 49.7 & 5.8  & 50.9 & 16.6 & 50.9 & 18.5 & 54.7 \\[1.5pt]
MV Calcification              &  8.6 & 45.4 & \textbf{72.2} & 44.5 & \underline{69.7} & 15.4 & 51.7 & 9.4  & 48.1 & 0.0  & 50.0 & 14.1 & 47.7 & 15.9 & 51.2 \\[1.5pt]
LV Enlargement                &  5.7 & 57.4 & \textbf{75.5} & 53.9 & \underline{71.5} & 11.2 & 52.5 & 8.8  & 47.9 & 0.0  & 50.0 & 11.1 & 51.1 & 5.0  & 46.2 \\[1.5pt]
RA Enlargement                &  4.8 & 25.9 & \textbf{58.7} & 18.4 & \underline{55.9} & 6.0  & 46.9 & 9.1  & 50.9 & 0.0  & 50.0 & 8.9  & 48.4 & 8.7  & 48.0 \\[1.5pt]
RV Enlargement                &  2.5 & 13.8 & \textbf{55.5} & 3.6  & 50.6 & 2.5  & 44.2 & 2.3  & 41.1 & 0.0  & 50.0 & 5.4  & \underline{53.8} & 5.2  & 52.1 \\[1.5pt]
Bicuspid AV                   &  2.5 & 39.2 & \textbf{65.7} & 19.2 & \underline{57.4} & 4.5  & 50.1 & 6.7  & 55.6 & 0.0  & 50.0 & 5.2  & 51.9 & 5.1  & 51.4 \\[1.5pt]
\rowcolor{yellow!12}
\textbf{Macro Average}        &      & \textbf{49.4} & \textbf{67.4} & \underline{45.1} & \underline{65.1} & 19.6 & 50.3 & 16.2 & 49.2 & 2.7  & 50.1 & 23.9 & 50.3 & 19.9 & 50.3 \\
\midrule
\multicolumn{16}{l}{\cellcolor{yellow!8}\textit{\textbf{MIMICEchoQA}~\cite{thapa2025mimicechoqa} (single-view)}} \\[3pt]
LA Enlargement, $n = 18$ & 77.8 & 11.6 & 52.7 & 6.1  & 45.8 & 58.3 & 37.5 & 66.7 & 32.1 & 0.0  & 50.0 & 66.7 & \underline{53.6} & 86.7 & \textbf{58.9} \\[1.5pt]
AV Regurgitation,  $n = 40$ & 67.5 & 38.3 & \textbf{64.0} & 31.1 & \underline{60.3} & 53.3 & 49.1 & 0.0  & 50.0 & 80.6 & 50.0 & 65.5 & 44.9 & 38.1 & 37.9 \\[1.5pt]
LV Systolic Dysfunction, $n = 28$ & 60.7 & 23.8 & \underline{60.1} & 22.7 & 59.7 & 30.0 & 58.8 & 30.0 & 58.8 & 0.0  & 50.0 & 78.9 & \textbf{66.8} & 62.9 & 50.5 \\[1.5pt]
RV Enlargement, $n = 17$ & 58.8 & 18.2 & \textbf{60.6} & 8.3  & \underline{52.1} & 52.6 & 46.4 & 18.2 & 55.0 & 0.0  & 50.0 & 72.0 & 52.1 & 44.4 & 41.4 \\[1.5pt]
AV Stenosis, $n = 52$ & 21.1 & 20.0 & \underline{57.5} & 31.6 & \textbf{62.5} & 27.6 & 51.1 & 0.0  & 50.0 & 0.0  & 47.6 & 27.0 & 47.1 & 28.6 & 55.1 \\[1.5pt]
\rowcolor{yellow!12}
\textbf{Macro Average}        &      & 22.4 & \textbf{59.0} & 20.0 & \underline{56.1} & 44.4 & 48.6 & 23.0 & 49.2 & 16.1 & 49.5 & \textbf{62.0} & 52.9 &  \underline{52.1} & 48.8 \\
\midrule
\multicolumn{16}{l}{\cellcolor{yellow!8}\textit{\textbf{EchoNet-Dynamic}~\cite{ouyang2020echonetdynamic} ($n = 1{,}277$, single-view)}} \\[3pt]
LV Systolic Dysfunction       &      & 39.3  & \textbf{60.2}  & 37.2 & \underline{58.7} & 5.6  & 50.6 & 51.2 & 50.5 & 66.7 & 50.0 & 36.0 & 51.8 & 7.7  & 50.0 \\
\bottomrule
\end{tabular}%
}
\end{table*}

\subsection{Results of Abnormality Classification}
\textbf{Private test set.}~\cref{tab:classification} presents per-disease classification results on the private multi-view test set. EchoSonar-R$^{\dagger}$ demonstrates strong abnormality classification performance, outperforming all baselines across both F1 and BAcc. For macro BAcc, EchoSonar-R$^{\dagger}$ achieves 67.4\%, surpassing the SFT-only EchoSonar-R$^{\ast}$ by 2.3\% and the strongest external baselines (Chiron-o1 and Lingshu) by over 17\%. For macro F1, EchoSonar-R$^{\dagger}$ achieves 49.4\%, approximately twice the best external baseline (Chiron-o1, 23.9\%). The advantage is particularly pronounced on low-prevalence conditions: for aortic stenosis (9.5\%), EchoSonar-R$^{\dagger}$ attains 82.2\% BAcc compared to 54.7\% for the strongest baseline; for LV enlargement (5.7\%), 75.5\% versus 52.5\%; and for bicuspid aortic valve (2.5\%), 65.7\% versus 55.6\%. Notably, most external baselines hover around 50\% BAcc across conditions, suggesting a drift toward majority-class predictions, resulting in reasonable F1 on common diseases while demonstrating poor separation between positive and negative cases. Comparing EchoSonar-R$^{\dagger}$ and EchoSonar-R$^{\ast}$, EchoSonar-R$^{\dagger}$ provides consistent improvements over SFT-only, with macro F1 rising from 45.1\% to 49.4\% and BAcc from 65.1\% to 67.4\%. The gains span 11 of 12 categories and are most pronounced for low-prevalence conditions, suggesting that reward-based optimization is particularly effective where SFT has a limited supervision signal. To assess statistical significance, we perform a per-class sign test: EchoSonar-R$^{\dagger}$ outperforms EchoSonar-R$^{\ast}$ on F1 in 11 of 12 categories and on BAcc in 11 of 12, yielding a sign-test $p \approx 0.006$, confirming that the GRPO gains are statistically significant. This pattern suggests that GRPO refines the model's decision boundaries rather than learning fundamentally new visual features. For the MV Regurgitation where EchoSonar-R$^{\ast}$ outperforms EchoSonar-R$^{\dagger}$, the difference is modest ($<$1\% BAcc), indicating that EchoSonar-R$^{\dagger}$ does not degrade performance on well-learned categories while substantially improving others.

\noindent \textbf{Public test sets.} On MIMICEchoQA in~\cref{tab:classification}, EchoSonar-R$^{\dagger}$ achieves the highest macro BAcc of 59.0\%, outperforming the Chiron-o1 by 6.1\%. While baselines such as Chiron-o1 and Lingshu achieve higher macro F1, their BAcc remains at or below 52.9\%. EchoSonar-R$^{\dagger}$ maintains the highest BAcc on 3 of 5 conditions, being robust to cases containing single-view inputs. For aortic stenosis, EchoSonar-R$^{\ast}$ achieves the highest BAcc of 62.5\%, outperforming Lingshu by 7.4\%. On EchoNet-Dynamic, EchoSonar-R$^{\dagger}$ achieves the highest BAcc of 60.2\% for LV systolic dysfunction, surpassing the Chiron-o1 by 8.4\%.

\noindent \textbf{Cross-dataset consistency.} Five conditions are shared between the private test set and MIMICEchoQA, allowing direct comparison of model robustness across dataset shift. EchoSonar-R$^{\dagger}$ exhibits the 8.4\% BAcc average drop across these conditions. AV regurgitation is the most stable for EchoSonar-R, while LA  enlargement shows the largest drop. This likely reflects both the difficulty of assessing chamber size from a single view and differences in diagnostic thresholds across institutions: trace or borderline findings may be reported as abnormal at one site but considered within normal limits at another. For LV systolic dysfunction, which can be evaluated across all three datasets, EchoSonar-R$^{\dagger}$ maintains strong performance: 74.0\% on the private set, 60.1\% on MIMICEchoQA, and 60.2\% on EchoNet-Dynamic. 
\begin{table*}[t]
\centering
\caption{Reasoning quality evaluation on the private test set, scored by Mistral-7B~\cite{jiang2023mistral} on a 1--5 Likert scale ($\uparrow$). EchoSonar-R$^{\dagger}$ denotes GRPO training; EchoSonar-R$^{\ast}$ denotes SFT-only. Best results are in \textbf{bold}, second best \underline{underlined}.}
\label{tab:reasoning}
\setlength{\tabcolsep}{4.5pt}
\resizebox{\textwidth}{!}{%
\begin{tabular}{l | c | c || c | c | c | c}
\toprule
\textbf{Metric} & \textbf{EchoSonar-R}$^{\dagger}$ & \textbf{EchoSonar-R}$^{\ast}$ & \textbf{Qwen3-VL} & \textbf{MedGemma} & \textbf{Chiron-o1} & \textbf{Lingshu} \\
\midrule
Reasoning-Answer Agreement               & \textbf{4.98} & \underline{4.96} & 3.29 & 3.33 & 3.50 & 4.02 \\[2pt]
Reasoning Efficiency    & \textbf{4.98} & \underline{4.97} & 3.73 & 3.74 & 3.80 & 4.58 \\[2pt]
Factual Correctness     & \underline{4.99} & \textbf{5.00} & 4.77 & 4.15 & 3.49 & 4.29 \\[2pt]
Evidence Grounding      & \textbf{5.00} & \textbf{5.00} & 3.31 & 3.71 & 3.56 & \underline{4.68} \\[2pt]
Terminology Accuracy    & \textbf{5.00} & \textbf{5.00} & \underline{4.96} & 4.29 & 3.69 & 4.70 \\[2pt]
\midrule
\rowcolor{yellow!12}
\textbf{Average}        & \textbf{4.99} & \textbf{4.99} & 4.01 & 3.84 & 3.61 & \underline{4.45} \\
\bottomrule
\end{tabular}%
}
\end{table*}

\noindent \textbf{Reasoning quality.}~\cref{tab:reasoning} reports reasoning trace quality across five dimensions. EchoSonar-R$^{\dagger}$ achieves near-perfect average scores (4.99/5.00), substantially outperforming all baselines. The gap is widest on Evidence Grounding, indicating that baseline reasoning traces rely on generic medical statements rather than referencing specific visual observations from the echocardiographic input. Reasoning--Answer Agreement shows a similar pattern: EchoSonar-R$^{\dagger}$ maintains near-perfect consistency between its reasoning and final answers (4.98), while Qwen3-VL (3.29) and MedGemma (3.33) frequently produce conclusions that contradict or omit findings discussed in their reasoning. Among baselines, Lingshu achieves the highest average (4.45), driven by strong Terminology Accuracy (4.70) and Evidence Grounding (4.68), yet its classification performance remains near chance (\cref{tab:classification}). 

\begin{table*}[t]
\centering
\caption{Report generation quality on the private test set. NLG 
metrics measure lexical and semantic overlap with reference reports 
($\uparrow$). GREEN~\cite{ostmeier2024green} evaluates clinical 
faithfulness by identifying and categorizing clinically significant 
errors in generated reports. EchoSonar-R$^{\dagger}$ denotes GRPO 
training; EchoSonar-R$^{\ast}$ denotes SFT-only. Best results are 
in \textbf{bold}, second best \underline{underlined}.}
\label{tab:report_generation}
\setlength{\tabcolsep}{4.5pt}
\resizebox{\textwidth}{!}{%
\begin{tabular}{l | c | c || c | c | c | c | c}
\toprule
\textbf{Metric} & \textbf{EchoSonar-R}$^{\dagger}$ & \textbf{EchoSonar-R}$^{\ast}$ & \textbf{Qwen3-VL} & \textbf{MedGemma} & \textbf{EchoPrime} & \textbf{Chiron-o1} & \textbf{Lingshu} \\
\midrule
\multicolumn{8}{l}{\cellcolor{yellow!8}\textit{\textbf{NLG Metrics} ($\uparrow$)}} \\[3pt]
BLEU-1                  & \textbf{0.795} & \underline{0.792} & 0.076 & 0.135 & 0.143 & 0.125 & 0.099 \\[1.5pt]
BLEU-2                  & \textbf{0.767} & \underline{0.762} & 0.027 & 0.066 & 0.061 & 0.039 & 0.031 \\[1.5pt]
BLEU-3                  & \textbf{0.740} & \underline{0.736} & 0.014 & 0.047 & 0.036 & 0.023 & 0.016 \\[1.5pt]
BLEU-4                  & \textbf{0.725} & \underline{0.720} & 0.010 & 0.038 & 0.027 & 0.019 & 0.011 \\[1.5pt]
METEOR                  & \textbf{0.829} & \underline{0.826} & 0.195 & 0.260 & 0.234 & 0.197 & 0.230 \\[1.5pt]
ROUGE-L                 & \textbf{0.819} & \underline{0.815} & 0.113 & 0.188 & 0.192 & 0.185 & 0.162 \\[1.5pt]
BERTScore               & \textbf{0.985} & \textbf{0.985} & 0.924 & 0.929 & \underline{0.933} & 0.931 & 0.927 \\[1.5pt]
\midrule
\multicolumn{8}{l}{\cellcolor{yellow!8}\textit{\textbf{GREEN Clinical Faithfulness}}} \\[3pt]
GREEN Score ($\uparrow$)        & \textbf{0.800} & \underline{0.796} & 0.216 & 0.453 & 0.306 & 0.358 & 0.491 \\[1.5pt]
Mean Sig.\ Errors ($\downarrow$) & \textbf{0.590} & \underline{0.602} & 1.747 & 1.275 & 1.588 & 1.502 & 1.193 \\[1.5pt]
\quad Hallucination ($\downarrow$)     & \textbf{0.175} & \underline{0.179} & 0.706 & 0.260 & 0.524 & 0.475 & 0.321 \\[1.5pt]
\quad Omission ($\downarrow$)          & \textbf{0.193} & \underline{0.198} & 0.677 & 0.716 & 0.585 & 0.613 & 0.633 \\[1.5pt]
\quad Wrong Location ($\downarrow$)    & \textbf{0.018} & \underline{0.019} & 0.044 & 0.027 & 0.047 & 0.044 & 0.015 \\
\bottomrule
\end{tabular}%
}
\end{table*}
\subsection{Results of Report Generation}
\cref{tab:report_generation} presents report generation quality on the private test set. EchoSonar-R$^{\dagger}$ achieves the highest scores across all NLG metrics by a wide margin. For BLEU-4, it reaches 0.725 compared to 0.010 for the Qwen3-VL, reflecting the advantage of training on reports from the same institutional format. Baselines that have never seen reports from this institution cannot match its lexical structure, making direct NLG comparisons with external models inherently asymmetric. BERTScore partially mitigates this by measuring semantic rather than surface-level overlap, yet EchoSonar-R$^{\dagger}$ still leads (0.985 vs.\ 0.933) for EchoPrime and 0.931 for Chiron-o1, suggesting that the advantage extends beyond formatting to clinical content.

The GREEN evaluation provides a more meaningful comparison, as it assesses clinical faithfulness independently of reporting style. EchoSonar-R$^{\dagger}$ achieves a GREEN score of 0.800, substantially above Lingshu (0.491) and Chiron-o1 (0.358). The error breakdown reveals that Qwen3-VL produces 1.747 mean significant errors per report, with omissions (0.677) and hallucinations (0.706). It frequently fabricates findings not present in the study or omits pathology. Lingshu, on the other hand, has the highest omission rate (0.633) but comparatively lower hallucination (0.321), suggesting it tends toward underreporting rather than fabrication.EchoSonar-R$^{\dagger}$ maintains the lowest error rates across all categories, with hallucination at 0.175 and omission at 0.193, and achieves the fewest wrong location errors (0.018).

Comparing EchoSonar-R$^{\dagger}$ and EchoSonar-R$^{\ast}$ variants, the differences are modest: GREEN score improves marginally from 0.796 to 0.800, with small reductions in hallucination and omission. This is expected, as the GRPO reward for report generation uses cosine similarity, which optimizes for semantic content coverage rather than fine-grained error reduction. The SFT stage, trained directly on reference reports, already captures most of the institutional reporting conventions, leaving limited room for GRPO to improve further on this task.

\subsection{Ablation Analysis}
\begin{table*}[t]
\centering
\caption{Ablation studies on the private test set. Top: effect of reasoning-augmented training data during SFT. Bottom: contribution of visual encoder components at inference. VT = video tokens, ST = structure tokens. Best results per section are in \textbf{bold}.}
\label{tab:ablation}
\setlength{\tabcolsep}{4pt}
\resizebox{0.6\textwidth}{!}{%
\begin{tabular}{l | cc | cccc}
\toprule
& & & \multicolumn{4}{c}{\textbf{Macro Metrics}} \\
\cmidrule(lr){4-7}
\textbf{Configuration} & \textbf{VT} & \textbf{ST} & \textbf{F1} & \textbf{BAcc} & \textbf{Sens.} & \textbf{Spec.} \\
\midrule
\multicolumn{7}{l}{\cellcolor{yellow!8}\textit{\textbf{Reasoning in Training}}} \\[3pt]
w/ reasoning targets          & \cmark & \cmark & \textbf{45.1} & 65.1 & 43.4 & \textbf{86.8} \\[1.5pt]
w/o reasoning targets         & \cmark & \cmark & 44.4 & \textbf{65.5} & \textbf{47.6} & 83.4 \\
\midrule
\multicolumn{7}{l}{\cellcolor{yellow!8}\textit{\textbf{Visual Input Components at Inference}}} \\[3pt]
Full                            & \cmark & \cmark & \textbf{45.1} & \textbf{65.1} & \textbf{43.4} & 86.8 \\[1.5pt]
Remove structure tokens         & \cmark & \xmark & 43.9 & 63.9 & 41.9 & 85.8 \\[1.5pt]
Remove video tokens             & \xmark & \cmark & 10.5 & 49.8 & 11.5 & 88.0 \\[1.5pt]
Remove both                     & \xmark & \xmark & 10.2 & 50.2 & 10.3 & \textbf{90.2} \\
\bottomrule
\end{tabular}%
}
\end{table*}

\textbf{Effect of reasoning in training.} \cref{tab:ablation} compares EchoSonar-R$^{\ast}$ trained with and without reasoning targets. Aggregate performance is comparable (macro F1: 45.1\% vs.\ 44.4\%; BAcc: 65.1\% vs.\ 65.5\%), suggesting that reasoning supervision does not substantially affect overall discriminative accuracy at SFT stage. The main change is in the sensitivity–specificity trade-off: removing reasoning traces increases sensitivity (47.6\% vs.\ 43.4\%) at the expense of specificity (83.4\% vs.\ 86.8\%). Thus, the reasoning-trained model behaves more conservatively, producing fewer positive predictions and fewer false positives, while the non-reasoning model identifies more true positives, but with more false positives. However, the reasoning traces support clinical transparency: they enable clinicians to audit the model’s stated rationale and more easily localize systematic failure modes than is possible with binary outputs alone.

\noindent \textbf{Visual input components at inference.} \cref{tab:ablation} evaluates the contribution of each token type by removing them from the input sequence $X$ at inference. The video tokens are the primary information source: removing them causes a significant drop in F1 (45.1\% to 10.5\%) and BAcc (65.1\% to 49.8\%), with the model falling to near-random discrimination. Removing both token types confirms this, with performance collapsing further to 10.2\% F1 and 50.2\% BAcc. Notably, the marginal difference between removing video tokens alone and removing both indicates that structure tokens carry negligible diagnostic signal without the accompanying video context. The structure tokens provide a complementary but smaller contribution: removing them while retaining video tokens produces a modest decrease in both F1 (45.1\% to 43.9\%) and BAcc (65.1\% to 63.9\%), confirming that the detector embeddings help calibrate predictions and improve discrimination between positive and negative cases. The rising specificity as tokens 
are removed (86.8\% to 88.0\% to 90.2\%) reflects a 
progressively more conservative model that defaults to 
predicting the negative class when deprived of visual evidence.

\begin{figure}[t]
    \centering
    \includegraphics[width=\linewidth]{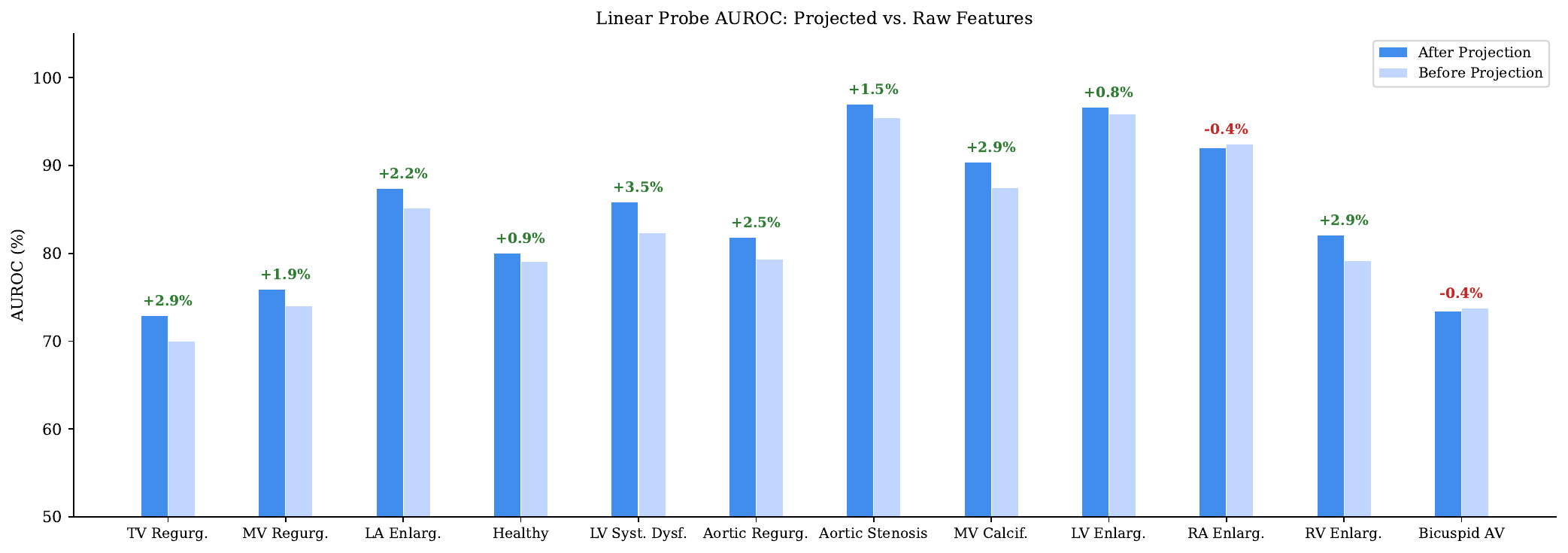}
    \caption{Linear probe AUROC before and after cross-modal 
    projection.}
    \label{fig:ablation_auroc}  
\end{figure}

\noindent \textbf{Effect of cross-modal projection.} To test whether the learned projectors $f_v$ enhance the clinical separability of visual features, we train linear probes on EchoPrime embeddings before and after projection and report per-disease AUROC in \cref{fig:ablation_auroc}. Projection improves AUROC for 10 of 12 conditions, with the largest gains for LV systolic dysfunction (3.5\%), RV enlargement (2.9\%), TV regurgitation (2.9\%), and MV calcification (2.9\%). The remaining two conditions, RA enlargement and bicuspid AV, show only minor decreases. Overall, these results indicate that $f_v$ reshapes the frozen encoder’s representation space to amplify the disease-relevant signal, acting as a learned task-specific adapter.

\subsection{Qualitative Results}
\begin{figure*}[t]
\centering
\scriptsize
\setlength{\tabcolsep}{3pt}
\renewcommand{\arraystretch}{1.05}
\resizebox{\textwidth}{!}{%
\begin{tabular}{@{}l p{0.46\textwidth} p{0.46\textwidth}@{}}
\multicolumn{3}{l}{\textbf{(a) Classification: Normal study, ground truth = no abnormalities}} \\
\toprule
& \textbf{EchoSonar-R$^{\dagger}$}: 0/11 errors, score: 5.0
& \textbf{Lingshu}: 10/11 errors, score: 4.2 \\
\midrule
\textit{R.} &
\textcolor{teal}{[PLAX]} LV normal. AV normal morphology, no calcification. MV normal mobility. 
\textcolor{teal}{[A4C]} LA within normal limits, \textcolor{blue}{consistent with PLAX}. RA normal. RV-to-LV ratio normal. 
\textcolor{teal}{[A4C Zoom]} Wall motion normal, \textcolor{blue}{confirming systolic function from PLAX}. 
\textcolor{teal}{[A2C]} Wall motion normal. LA \textcolor{blue}{normal across views}. 
\textcolor{teal}{[PSAX]} No PR jet. 
\textcolor{teal}{[A3C]} AV, MV \textcolor{blue}{confirmed normal across all views}. &
\textcolor{red}{Turbulent flow across TV, indicating regurgitation}. \textcolor{red}{MR as seen by backflow into LA}. AV appears \textcolor{red}{thickened and calcified}. MV leaflets show \textcolor{red}{calcification}. RA appears \textcolor{red}{enlarged}. LV shows \textcolor{red}{reduced contractility}. \\
\textit{A.} &
No significant abnormalities. \cmark &
TR, MR, AS, MV Calc., RAE, RVE, LAE, LVSD, LVE \,\xmark \\
\midrule
& \textbf{EchoSonar-R$^{\ast}$} (SFT): 1/11 errors, score: 5.0
& \textbf{Chiron-o1}: 11/11 errors, score: 1.4 \\
\midrule
\textit{R.} &
\textcolor{teal}{[A4C Zoom]} LV normal. \textcolor{orange}{LA enlarged, likely diastolic dysfunction}. RV normal. 
\textcolor{teal}{[A2C]} LA enlarged. 
\textcolor{teal}{[PLAX]} LV, MV normal. AV trileaflet. &
\textcolor{gray}{Based on the analysis of the echocardiographic images, the identified abnormalities are: TR: Present. MR: Present. AS: Present. MV Calc.: Present. RAE: Present. RVE: Present. LAE: Present. LVSD: Present. LVE: Present. BAV: Present. AR: Present.} \\
\textit{A.} &
Left Atrial Enlargement \,\xmark\,{\tiny(FP)} &
All 11 abnormalities ``Present'' \,\xmark \\
\midrule
\multicolumn{3}{l}{\textbf{(b) Report generation: Per-section comparison}} \\
\midrule
\end{tabular}%
}

\setlength{\tabcolsep}{4pt}
\renewcommand{\arraystretch}{1.2}
\scriptsize
\resizebox{\textwidth}{!}{%
\begin{tabular}{@{}p{0.07\textwidth} p{0.22\textwidth} p{0.22\textwidth} p{0.22\textwidth} p{0.22\textwidth}@{}}
\textbf{Section} & \textbf{Reference} & 
\textbf{EchoSonar-R$^{\dagger}$} & 
\textbf{Chiron-o1} & \textbf{Lingshu} \\
\midrule
LV &
Mildly reduced systolic function. Mild concentric LVH. &
\textcolor{teal}{Mild concentric LVH.} \textcolor{orange}{Normal systolic function.} &
\textcolor{orange}{Normal size and function.} &
\textcolor{red}{Normal, no hypertrophy.} \\
MV &
Normal appearance. Mild MR. &
Normal appearance. \textcolor{orange}{Trace MR.} &
\textcolor{red}{Normal, no regurgitation.} &
\textcolor{red}{Normal, no regurgitation.} \\
Concl. &
Mild hypokinesis. Mild LVH. Mild LAE. Mild MR. &
\textcolor{teal}{Mild LVH.} \textcolor{orange}{Normal EF.} &
\textcolor{red}{Normal, no valvular disease.} &
\textcolor{red}{Normal, no abnormalities.} \\
\bottomrule
\end{tabular}%
}
\caption{Qualitative examples. 
\textcolor{teal}{correct/grounded}, 
\textcolor{blue}{cross-view confirmation}, 
\textcolor{orange}{severity error}, 
\textcolor{red}{fabricated/missed}. }
\label{fig:qualitative}
\end{figure*}

\cref{fig:qualitative} illustrates representative model outputs. Panel~(a) shows a normal study in which EchoSonar-R$^{\dagger}$ correctly identifies the absence of disease by systematically reviewing each cardiac structure across five views and explicitly cross-checking consistency across views (e.g., verifying that normal LA dimensions in A4C agree with the PLAX assessment). Lingshu attains an average reasoning score of 4.2, yet it hallucinates pathologies with convincing clinical language, describing ``turbulent flow across the tricuspid valve'' and ``thickened and calcified'' aortic leaflets in a structurally normal heart. Chiron-o1, on the other hand, outputs a flat list labeling all 11 conditions as present with no reasoning or visual evidence. The EchoSonar-R$^{\ast}$ variant produces one false positive, which EchoSonar-R$^{\dagger}$ learns to suppress as diastolic dysfunction.

Panel~(b) presents a more challenging case with multiple mild abnormalities present at different sections of the report. EchoSonar-R$^{\dagger}$ correctly identifies mild concentric LVH but underestimates the severity of systolic dysfunction. In contrast, Chiron-o1 and Lingshu both report entirely normal findings across all sections, with no evidence of LVH, LA enlargement, or mitral regurgitation. This pattern is consistent with their near-chance BAcc in \cref{tab:classification} and high omission rates in \cref{tab:report_generation}.

\label{sec:qualitative_results}

\section{Conclusion}
In this paper, we present EchoSonar-R, a reasoning-enabled multi-view vision--language model for echocardiographic disease classification and report generation. By integrating a spatiotemporal video encoder with a structure-aware cardiac detector, our model provides the language model with both global motion dynamics and spatially grounded anatomical cues across the full echocardiographic study. A two-stage training strategy enables the model to produce reasoning while optimizing directly for clinical correctness. Our extensive experiments across a private multi-view dataset and two public benchmarks demonstrate strong performance on both abnormality classification and report generation, improving macro 
balanced accuracy by +17.1\% and +6.1\% over the strongest baselines on private and public sets, respectively, while halving the mean clinically significant errors in generated reports (0.590 vs.\ 1.193), with ablations and qualitative analysis validating each component.

\noindent\textbf{Limitations and Future Work.} 
The current model processes each study independently without leveraging prior examinations from the same patient. We plan to expand EchoSonar-R to incorporate longitudinal patient history, enabling tracking of disease progression across serial echocardiograms.

\section*{Acknowledgements}
The authors would like to thank Roman Sendler and Joseph Sokol from iCardio for their valuable feedback and insight.

%
%
\bibliographystyle{splncs04}
\bibliography{main}

@String(CVPR  = {IEEE Conf. Comput. Vis. Pattern Recog.})

@String(ECCV  = {Eur. Conf. Comput. Vis.})

@String(NeurIPS = {Adv. Neural Inform. Process. Syst.})

@String(ICML  = {Int. Conf. Mach. Learn.})

@String(ICLR  = {Int. Conf. Learn. Represent.})

@String(CVPR  = {CVPR})

@String(ECCV  = {ECCV})

@String(NeurIPS = {NeurIPS})

@String(ICML  = {ICML})

@String(ICLR  = {ICLR})

@misc{who2024cvd,
  author       = {{World Health Organization}},
  title        = {Cardiovascular Diseases ({CVDs}): Fact Sheet},
  year         = {2024},
  howpublished = {\url{https://www.who.int/news-room/fact-sheets/detail/cardiovascular-diseases-(cvds)}},
  note         = {Accessed: 2025-05-01}
}

@article{roth2024heartofworld,
  author    = {Roth, Gregory A. and Mensah, George A. and Fuster, Valentin},
  title     = {The Global Burden of Cardiovascular Diseases and Risks: A Compass for Global Action},
  journal   = {Journal of the American College of Cardiology},
  volume    = {82},
  number    = {25},
  pages     = {2350--2473},
  year      = {2023},
  doi       = {10.1016/j.jacc.2023.11.007}
}

@incollection{statpearls2023echoimaging,
  title={Echocardiography imaging techniques},
  author={Ahmed, Intisar and Sasikumar, Navaneetha},
  booktitle={StatPearls [Internet]},
  year={2023},
  publisher={StatPearls Publishing}
}

@article{chambers2012echofrontier,
  author    = {Chambers, John},
  title     = {Echocardiography: Frontier Imaging in Cardiology},
  journal   = {British Journal of Hospital Medicine},
  year      = {2012},
  note      = {PMC3473911}
}

@article{won2024sonographershortage,
  author    = {Won, Daniel and Walker, James and Horowitz, Russ and Bharadwaj, Sandeep and Carlton, Edward and Gabriel, Helena},
  title     = {Sound the Alarm: The Sonographer Shortage Is Echoing Across Healthcare},
  journal   = {Journal of Ultrasound in Medicine},
  volume    = {43},
  number    = {7},
  pages     = {1289--1301},
  year      = {2024},
  doi       = {10.1002/jum.16453}
}

@article{escatlas2024cvd,
  author  = {Timmis, Adam and Aboyans, Victor and Vardas,               Panos and
             Townsend, Nick and Torbica, Aleksandra and Kavousi, Maryam and
             Boriani, Giuseppe and Huculeci, Radu and Kazakiewicz, Denis and
             Scherr, Daniel and Karagiannidis, Efstratios and Cvijic, Marta and
             Kap{\l}on-Cie{\'s}licka, Agnieszka and Ignatiuk, Barbara and
             Raatikainen, Pekka and {De Smedt}, Delphine and Wood, Angela and
             Dudek, Dariusz and {Van Belle}, Eric and Weidinger, Franz and
             {ESC National Cardiac Societies}},
  title   = {European Society of Cardiology: the 2023 {Atlas} of Cardiovascular Disease Statistics},
  journal = {European Heart Journal},
  volume  = {45},
  number  = {38},
  pages   = {4019--4062},
  year    = {2024},
  month   = oct,
  doi     = {10.1093/eurheartj/ehae466},
  pmid    = {39189413}
}

@article{arega2023imagingafrica,
  author  = {Lakshmanan, Suvasini and Mbanze, Irina},
  title   = {A comparison of cardiovascular imaging practices in {Africa}, {North America}, and {Europe}: two faces of the same coin},
  journal = {European Heart Journal - Imaging Methods and Practice},
  volume  = {1},
  number  = {1},
  pages   = {qyad005},
  year    = {2023},
  month   = jul,
  doi     = {10.1093/ehjimp/qyad005},
  pmid    = {39044787},
  pmcid   = {PMC11195774}
}

@article{vukadinovic2024echoprime,
  author    = {Vukadinovic, Milos and Xu, Alan and Cheng, Xiu and Kwan, Alan C. and Ouyang, David},
  title     = {{EchoPrime}: A Multi-Video View-Informed Vision-Language Model for Comprehensive Echocardiography Interpretation},
  journal   = {arXiv preprint arXiv:2410.09704},
  year      = {2024}
}

@article{christensen2024echoclip,
  title={Vision--language foundation model for echocardiogram interpretation},
  author={Christensen, Matthew and Vukadinovic, Milos and Yuan, Neal and Ouyang, David},
  journal={Nature Medicine},
  volume={30},
  number={5},
  pages={1481--1488},
  year={2024},
  publisher={Nature Publishing Group US New York}
}

@article{ghorbani2020deeplearningecho,
  author    = {Ghorbani, Amirata and Ouyang, David and Abid, Abubakar and He, Bryan and Chen, Jonathan H. and Harrington, Robert A. and Liang, David H. and Ashley, Euan A. and Zou, James Y.},
  title     = {Deep Learning Interpretation of Echocardiograms},
  journal   = {npj Digital Medicine},
  volume    = {3},
  pages     = {10},
  year      = {2020},
  doi       = {10.1038/s41746-019-0216-8}
}

@article{zhang2018fullyautomatedecho,
  title={Fully automated echocardiogram interpretation in clinical practice: feasibility and diagnostic accuracy},
  author={Zhang, Jeffrey and Gajjala, Sravani and Agrawal, Pulkit and Tison, Geoffrey H and Hallock, Laura A and Beussink-Nelson, Lauren and Lassen, Mats H and Fan, Eugene and Aras, Mandar A and Jordan, ChaRandle and others},
  journal={Circulation},
  volume={138},
  number={16},
  pages={1623--1635},
  year={2018},
  publisher={Lippincott Williams \& Wilkins Hagerstown, MD}
}

@article{savarese2023cvdhealthcare,
 author  = {Chong, Bryan and Jayabaskaran, Jayanth and      Jauhari, Silingga Metta and
             Chan, Siew Pang and Goh, Rachel and Kueh, Martin Tze Wah and Li, Henry and
             Chin, Yip Han and Kong, Gwyneth and Anand, Vickram Vijay and
             Wang, Jiong-Wei and Muthiah, Mark and Jain, Vardhmaan and Mehta, Anurag and
             Lim, Shir Lynn and Foo, Roger and Figtree, Gemma A and Nicholls, Stephen J and
             Mamas, Mamas A and Januzzi, James L and Chew, Nicholas W S and
             Richards, A Mark and Chan, Mark Y},
  title   = {Global burden of cardiovascular diseases: projections from 2025 to 2050},
  journal = {European Journal of Preventive Cardiology},
  volume  = {32},
  number  = {11},
  pages   = {1001--1015},
  year    = {2025},
  doi     = {10.1093/eurjpc/zwae281},
  pmid    = {39270739}
}

@article{jansen2024automated,
  title={Automated echocardiography view classification and quality assessment with recognition of unknown views},
  author={Jansen, Gino E and de Vos, Bob D and Molenaar, Mitchel A and Schuuring, Mark J and Bouma, Berto J and I{\v{s}}gum, Ivana},
  journal={Journal of Medical Imaging},
  volume={11},
  number={5},
  pages={054002--054002},
  year={2024},
  publisher={Society of Photo-Optical Instrumentation Engineers}
}

@article{maani2024simlvseg,
  title={SimLVSeg: simplifying left ventricular segmentation in 2-D+ time echocardiograms with self-and weakly supervised learning},
  author={Maani, Fadillah and Ukaye, Asim and Saadi, Nada and Saeed, Numan and Yaqub, Mohammad},
  journal={Ultrasound in Medicine \& Biology},
  volume={50},
  number={12},
  pages={1945--1954},
  year={2024},
  publisher={Elsevier}
}

@article{asch2019automated,
  title={Automated echocardiographic quantification of left ventricular ejection fraction without volume measurements using a machine learning algorithm mimicking a human expert},
  author={Asch, Federico M and Poilvert, Nicolas and Abraham, Theodore and Jankowski, Madeline and Cleve, Jayne and Adams, Michael and Romano, Nathanael and Hong, Ha and Mor-Avi, Victor and Martin, Randolph P and others},
  journal={Circulation: Cardiovascular Imaging},
  volume={12},
  number={9},
  pages={e009303},
  year={2019},
  publisher={Lippincott Williams \& Wilkins Hagerstown, MD}
}

@inproceedings{rahman2023deep,
  title={Deep learning-based left ventricular ejection fraction estimation from echocardiographic videos},
  author={Rahman, Shafiur and Haque, Rezaul and Swapno, SM Masfequier Rahman and Islam, Md Babul and Nobel, SM Nuruzzaman and others},
  booktitle={2023 International Conference on Evolutionary Algorithms and Soft Computing Techniques (EASCT)},
  pages={1--6},
  year={2023},
  organization={IEEE}
}

@article{li2023echoefnet,
  title={EchoEFNet: Multi-task deep learning network for automatic calculation of left ventricular ejection fraction in 2D echocardiography},
  author={Li, Honghe and Wang, Yonghuai and Qu, Mingjun and Cao, Peng and Feng, Chaolu and Yang, Jinzhu},
  journal={Computers in Biology and Medicine},
  volume={156},
  pages={106705},
  year={2023},
  publisher={Elsevier}
}

@article{holste2025panecho,
  title={PanEcho: Complete AI-enabled echocardiography interpretation with multi-task deep learning},
  author={Holste, Gregory and Oikonomou, Evangelos K and Tokodi, M{\'a}rton and Kov{\'a}cs, Attila and Wang, Zhangyang and Khera, Rohan},
  journal={medRxiv},
  pages={2024--11},
  year={2025}
}

@article{salih2023xaicardiacimaging,
  author    = {Salih, Ammar and Boscolo Galazzo, Ilaria and Raisi-Estabragh, Zahra and 
               Petersen, Steffen E. and Menegaz, Gloria and Salih, Amir},
  title     = {Explainable Artificial Intelligence and Cardiac Imaging: Toward More 
               Interpretable Models},
  journal   = {Circulation: Cardiovascular Imaging},
  volume    = {16},
  number    = {4},
  pages     = {e014519},
  year      = {2023},
  doi       = {10.1161/CIRCIMAGING.122.014519}
}

@inproceedings{vinyals2015showandtell,
  author    = {Vinyals, Oriol and Toshev, Alexander and Bengio, Samy and Erhan, Dumitru},
  title     = {Show and Tell: A Neural Image Caption Generator},
  booktitle = {Proceedings of the IEEE Conference on Computer Vision and Pattern 
               Recognition (CVPR)},
  pages     = {3156--3164},
  year      = {2015}
}

@inproceedings{xu2015showattendtell,
  author    = {Xu, Kelvin and Ba, Jimmy and Kiros, Ryan and Cho, Kyunghyun and 
               Courville, Aaron and Salakhutdinov, Ruslan and Zemel, Richard and Bengio, Yoshua},
  title     = {Show, Attend and Tell: Neural Image Caption Generation with Visual Attention},
  booktitle = {Proceedings of the 32nd International Conference on Machine Learning (ICML)},
  pages     = {2048--2057},
  year      = {2015}
}

@inproceedings{vaswani2017attention,
  author    = {Vaswani, Ashish and Shazeer, Noam and Parmar, Niki and Uszkoreit, Jakob and 
               Jones, Llion and Gomez, Aidan N. and Kaiser, {\L}ukasz and Polosukhin, Illia},
  title     = {Attention is All You Need},
  booktitle = {Advances in Neural Information Processing Systems (NeurIPS)},
  volume    = {30},
  year      = {2017}
}

@article{realenosei2024medicalcaptionsurvey,
  author    = {Reale-Nosei, Gabriel and Amador-Dom{\'\i}nguez, Elvira and Serrano, Emilio},
  title     = {From Vision to Text: A Comprehensive Review of Natural Image Captioning 
               in Medical Diagnosis and Radiology Report Generation},
  journal   = {Medical Image Analysis},
  volume    = {97},
  pages     = {103264},
  year      = {2024},
  doi       = {10.1016/j.media.2024.103264}
}

@inproceedings{chen2020r2gen,
  author    = {Chen, Zhihong and Song, Yan and Chang, Tsung-Hui and Wan, Xiang},
  title     = {Generating Radiology Reports via Memory-Driven Transformer},
  booktitle = {Proceedings of the 2020 Conference on Empirical Methods in Natural 
               Language Processing (EMNLP)},
  pages     = {1439--1449},
  year      = {2020}
}

@article{chen2022r2gengpt,
  author    = {Chen, Zhihong and Shen, Yaling and Song, Yan and Wan, Xiang},
  title     = {Cross-Modal Memory Networks for Radiology Report Generation},
  journal   = {arXiv preprint arXiv:2204.13258},
  year      = {2022}
}

@article{syryca2025echoreportllm,
  author    = {Syryca, Fabian and Gr{\"a}{\ss}er, Christian and Trenkwalder, Teresa and others},
  title     = {Automated Generation of Echocardiography Reports Using Artificial Intelligence: 
               A Novel Approach to Streamlining Cardiovascular Diagnostics},
  journal   = {The International Journal of Cardiovascular Imaging},
  volume    = {41},
  pages     = {967--977},
  year      = {2025},
  doi       = {10.1007/s10554-025-03382-1}
}

@article{chao2025echogpt,
  author    = {Chao, Chieh-Ju and Banerjee, Imon and Arsanjani, Reza and Ayoub, Chadi and 
               Tseng, Andrew and Delbrouck, Jean-Benoit and Kane, Garvan C. and 
               Lopez-Jimenez, Francisco and Attia, Zachi and Oh, Jae K. and 
               Erickson, Bradley and Fei-Fei, Li and Adeli, Ehsan and Langlotz, Curtis},
  title     = {Evaluating Large Language Models in Echocardiography Reporting: 
               Opportunities and Challenges},
  journal   = {European Heart Journal -- Digital Health},
  volume    = {6},
  number    = {3},
  pages     = {326--339},
  year      = {2025},
  doi       = {10.1093/ehjdh/ztae086}
}

@inproceedings{radford2021clip,
  author    = {Radford, Alec and Kim, Jong Wook and Hallacy, Chris and Ramesh, Aditya and 
               Goh, Gabriel and Agarwal, Sandhini and Sastry, Girish and Askell, Amanda and 
               Mishkin, Pamela and Clark, Jack and Krueger, Gretchen and Sutskever, Ilya},
  title     = {Learning Transferable Visual Models From Natural Language Supervision},
  booktitle = {Proceedings of the 38th International Conference on Machine Learning (ICML)},
  pages     = {8748--8763},
  year      = {2021}
}

@article{sellergren2025medgemma,
  author    = {Sellergren, Andrew and Kazemzadeh, Sahar and Jaroensri, Tiam and 
               Kiraly, Atilla and Traverse, Madeleine and Kohlberger, Timo and others},
  title     = {{MedGemma} Technical Report},
  journal   = {arXiv preprint arXiv:2507.05201},
  year      = {2025}
}

@article{xu2025lingshu,
  author    = {Xu, Weiwen and Chan, Hou Pong and Li, Long and Aljunied, Mahani and 
               Yuan, Ruifeng and Wang, Jianyu and Xiao, Chenghao and Chen, Guizhen and 
               Liu, Chaoqun and Li, Zhaodonghui and others},
  title     = {Lingshu: A Generalist Foundation Model for Unified Multimodal Medical 
               Understanding and Reasoning},
  journal   = {arXiv preprint arXiv:2506.07044},
  year      = {2025}
}

@inproceedings{wei2022cot,
  author    = {Wei, Jason and Wang, Xuezhi and Schuurmans, Dale and Bosma, Maarten and 
               Ichter, Brian and Xia, Fei and Chi, Ed and Le, Quoc V. and Zhou, Denny},
  title     = {Chain-of-Thought Prompting Elicits Reasoning in Large Language Models},
  booktitle = {Advances in Neural Information Processing Systems (NeurIPS)},
  volume    = {35},
  year      = {2022}
}

@article{pan2025medvlmr1,
  author    = {Pan, Jiazhen and Liu, Che and Wu, Junde and Liu, Fenglin and Zhu, Jiayuan and 
               Li, Hongwei Bran and Chen, Chen and Ouyang, Cheng and Rueckert, Daniel},
  title     = {{MedVLM-R1}: Incentivizing Medical Reasoning Capability of Vision-Language 
               Models via Reinforcement Learning},
  journal   = {arXiv preprint arXiv:2502.19634},
  year      = {2025}
}

@inproceedings{sun2025chirono1,
  author    = {Sun, Haoran and others},
  title     = {{Chiron-o1}: Igniting Multimodal Large Language Models towards Generalizable 
               Medical Reasoning via Mentor-Intern Collaborative Search},
  booktitle = {Advances in Neural Information Processing Systems (NeurIPS)},
  year      = {2025}
}

@article{she2025echovlm,
  author    = {She, Chaoyin and Lu, Ruifang and Chen, Lida and Wang, Wei and Huang, Qinghua},
  title     = {{EchoVLM}: Dynamic Mixture-of-Experts Vision-Language Model for Universal 
               Ultrasound Intelligence},
  journal   = {arXiv preprint arXiv:2509.14977},
  year      = {2025}
}

@inproceedings{yan2023contrastive,
  title={Style-aware radiology report generation with radgraph and few-shot prompting},
  author={Yan, Benjamin and Liu, Ruochen and Kuo, David and Adithan, Subathra and Reis, Eduardo and Kwak, Stephen and Venugopal, Vasantha and O’Connell, Chloe and Saenz, Agustina and Rajpurkar, Pranav and others},
  booktitle={Findings of the Association for Computational Linguistics: EMNLP 2023},
  pages={14676--14688},
  year={2023}
}

@article{li2023knowledgegraph,
  author    = {Li, Mingjie and Lin, Bingqian and Chen, Zicong and Lin, Haokun and 
               Liang, Xiaodan and Chang, Xiaojun},
  title     = {Auxiliary Signal-Guided Knowledge Encoder-Decoder for Medical Report Generation},
  journal   = {World Wide Web},
  volume    = {26},
  number    = {1},
  pages     = {253--270},
  year      = {2023},
  doi       = {10.1007/s11280-022-01013-6}
}

@inproceedings{zhao2024rtdetr,
  author    = {Zhao, Yian and Lv, Wenyu and Xu, Shangliang and Wei, Jinman and 
               Wang, Guanzhong and Dang, Qingqing and Liu, Yi and Chen, Jie},
  title     = {{DETRs} Beat {YOLOs} on Real-time Object Detection},
  booktitle = {Proceedings of the IEEE/CVF Conference on Computer Vision and 
               Pattern Recognition (CVPR)},
  pages     = {16965--16974},
  year      = {2024}
}

@inproceedings{teed2020raft,
  author    = {Teed, Zachary and Deng, Jia},
  title     = {{RAFT}: Recurrent All-Pairs Field Transforms for Optical Flow},
  booktitle = {European Conference on Computer Vision (ECCV)},
  pages     = {402--419},
  year      = {2020}
}

@inproceedings{papineni2002bleu,
  author    = {Papineni, Kishore and Roukos, Salim and Ward, Todd and Zhu, Wei-Jing},
  title     = {{BLEU}: A Method for Automatic Evaluation of Machine Translation},
  booktitle = {Proceedings of the 40th Annual Meeting of the Association for 
               Computational Linguistics (ACL)},
  pages     = {311--318},
  year      = {2002}
}

@inproceedings{lin2004rouge,
  author    = {Lin, Chin-Yew},
  title     = {{ROUGE}: A Package for Automatic Evaluation of Summaries},
  booktitle = {Text Summarization Branches Out},
  pages     = {74--81},
  year      = {2004}
}

@inproceedings{banerjee2005meteor,
  author    = {Banerjee, Satanjeev and Lavie, Alon},
  title     = {{METEOR}: An Automatic Metric for {MT} Evaluation with Improved 
               Correlation with Human Judgments},
  booktitle = {Proceedings of the ACL Workshop on Intrinsic and Extrinsic 
               Evaluation Measures for Machine Translation and/or Summarization},
  pages     = {65--72},
  year      = {2005}
}

@inproceedings{zhang2020bertscore,
  author    = {Zhang, Tianyi and Kishore, Varsha and Wu, Felix and Weinberger, 
               Kilian Q. and Artzi, Yoav},
  title     = {{BERTScore}: Evaluating Text Generation with {BERT}},
  booktitle = {International Conference on Learning Representations (ICLR)},
  year      = {2020}
}

@article{zheng2024judging,
  author    = {Zheng, Lianmin and Chiang, Wei-Lin and Sheng, Ying and Zhuang, 
               Siyuan and Wu, Zhanghao and Zhuang, Yonghao and Lin, Zi and 
               Li, Zhuohan and Li, Dacheng and Xing, Eric P. and others},
  title     = {Judging {LLM}-as-a-Judge with {MT-Bench} and {Chatbot Arena}},
  journal   = {Advances in Neural Information Processing Systems (NeurIPS)},
  volume    = {36},
  year      = {2024}
}

@misc{gow2023mimicivecho,
  title={Mimic-iv-echo: Echocardiogram matched subset},
  author={Gow, Brian and Pollard, Tom and Greenbaum, Nathaniel and Moody, Benjamin and Johnson, Alistair and Herbst, Elizabeth and Waks, Jonathan W and Eslami, Parastou and Chaudhari, Ashish and Carbonati, Tanner and others},
  journal={PhysioNet https://doi. org/10.13026/EF48-V217},
  year={2023}
}

@article{physionet2000,
  author    = {Goldberger, Ary L. and Amaral, Luis A. N. and Glass, Leon and 
               Hausdorff, Jeffrey M. and Ivanov, Plamen Ch. and Mark, Roger G. and 
               Mietus, Joseph E. and Moody, George B. and Peng, Chung-Kang and 
               Stanley, H. Eugene},
  title     = {{PhysioBank, PhysioToolkit, and PhysioNet}: Components of a New 
               Research Resource for Complex Physiologic Signals},
  journal   = {Circulation},
  volume    = {101},
  number    = {23},
  pages     = {e215--e220},
  year      = {2000}
}

@misc{thapa2025mimicechoqa,
  title={Mimic-iv-echo-ext-mimicechoqa: A benchmark dataset for echocardiogram-based visual question answering},
  author={Thapa, Rahul and Li, Andrew and Wu, Qingyang and He, Bryan and Sahashi, Yuki and Binder-Rodriguez, Christina and Zhang, Angela and Ouyang, David and Zou, James},
  journal={PhysioNet https://doi. org/10.13026/rndk-4s36. Version},
  volume={1},
  number={0},
  year={2025}
}

@article{lang2015ase,
  author    = {Lang, Roberto M. and Badano, Luigi P. and Mor-Avi, Victor and 
               Afilalo, Jonathan and Armstrong, Anderson and Ernande, Laura and 
               Flachskampf, Frank A. and Foster, Elyse and Goldstein, Steven A. 
               and Kuznetsova, Tatiana and Lancellotti, Patrizio and Muraru, 
               Denisa and Picard, Michael H. and Rietzschel, Ernst R. and 
               Rudski, Lawrence and Spencer, Kirk T. and Tsang, Wendy and 
               Voigt, J{\"o}rg-Uwe},
  title     = {Recommendations for Cardiac Chamber Quantification by 
               Echocardiography in Adults: An Update from the {American Society 
               of Echocardiography} and the {European Association of 
               Cardiovascular Imaging}},
  journal   = {European Heart Journal -- Cardiovascular Imaging},
  volume    = {16},
  number    = {3},
  pages     = {233--271},
  year      = {2015}
}

@article{ouyang2020echonetdynamic,
  author    = {Ouyang, David and He, Bryan and Ghorbani, Amirata and 
               Yuan, Neal and Ebinger, Joseph and Langlotz, Curt P. and 
               Heidenreich, Paul A. and Harrington, Robert A. and 
               Liang, David H. and Ashley, Euan A. and Zou, James Y.},
  title     = {Video-based {AI} for beat-to-beat assessment of cardiac function},
  journal   = {Nature},
  volume    = {580},
  number    = {7802},
  pages     = {252--256},
  year      = {2020},
  doi       = {10.1038/s41586-020-2145-8}
}

@inproceedings{ostmeier2024green,
  author    = {Ostmeier, Sophie and Xu, Justin and Chen, Zhihong and 
               Varma, Maya and Blankemeier, Louis and Bluethgen, Christian and 
               Michalson, Arne Edward and Moseley, Michael and Langlotz, Curtis 
               and Chaudhari, Akshay S. and Delbrouck, Jean-Benoit},
  title     = {{GREEN}: Generative Radiology Report Evaluation and Error 
               Notation},
  booktitle = {Findings of the Association for Computational Linguistics: 
               EMNLP 2024},
  pages     = {374--390},
  year      = {2024},
  doi       = {10.18653/v1/2024.findings-emnlp.21}
}

@article{jiang2023mistral,
  author    = {Jiang, Albert Q. and Sablayrolles, Alexandre and Mensch, Arthur 
               and Bamford, Chris and Chaplot, Devendra Singh and 
               de las Casas, Diego and Bressand, Florian and Lengyel, Gianna 
               and Lample, Guillaume and Saulnier, Lucile and 
               Lavaud, L{\'e}lio Renard and Lachaux, Marie-Anne and 
               Stock, Pierre and Le Scao, Teven and Lavril, Thibaut and 
               Wang, Thomas and Lacroix, Timoth{\'e}e and 
               El Sayed, William},
  title     = {Mistral {7B}},
  journal   = {arXiv preprint arXiv:2310.06825},
  year      = {2023}
}

@article{shao2024deepseekmath,
  author    = {Shao, Zhihong and Wang, Peiyi and Zhu, Qihao and Xu, Runxin 
               and Song, Junxiao and Zhang, Mingchuan and Li, Y.K. and Wu, Y. 
               and Guo, Daya},
  title     = {{DeepSeekMath}: Pushing the Limits of Mathematical Reasoning 
               in Open Language Models},
  journal   = {arXiv preprint arXiv:2402.03300},
  year      = {2024}
}

@article{ouyang2022training,
  author    = {Ouyang, Long and Wu, Jeffrey and Jiang, Xu and 
               Almeida, Diogo and Wainwright, Carroll and Mishkin, 
               Pamela and Zhang, Chong and Agarwal, Sandhini and 
               Slama, Katarina and Ray, Alex and others},
  title     = {Training language models to follow instructions 
               with human feedback},
  journal   = {Advances in Neural Information Processing Systems},
  volume    = {35},
  pages     = {27730--27744},
  year      = {2022}
}

@article{yu2025dapo,
  title={DAPO: An Open-Source LLM Reinforcement Learning System 
         that Goes Beyond},
  author={Yu, Qiying and others},
  journal={arXiv preprint arXiv:2503.14476},
  year={2025}
}

@article{liu2025understanding,
  title={Understanding R1-Zero-Like Training: A Critical Perspective},
  author={Liu, Zichen and others},
  journal={arXiv preprint arXiv:2503.20783},
  year={2025}
}

@article{yang2025qwen3,
  title={Qwen3 Technical Report},
  author={Yang, An and Yang, Baosong and Zhang, Beichen and 
          Wang, Binyuan and others},
  journal={arXiv preprint arXiv:2505.09388},
  year={2025}
}

@article{mitchell2019guidelines,
  title={Guidelines for Performing a Comprehensive Transthoracic 
         Echocardiographic Examination in Adults: Recommendations 
         from the {American Society of Echocardiography}},
  author={Mitchell, Cathy and Rahko, Peter S and Blauwet, Lori A 
          and Canaday, Brian and Finstuen, Jeanne A and Foster, 
          Maryellen C and Horton, Karen and Ogunyankin, Kofo O 
          and Palma, Richard A and Velazquez, Eric J},
  journal={Journal of the American Society of Echocardiography},
  volume={32},
  number={1},
  pages={1--64},
  year={2019},
  doi={10.1016/j.echo.2018.06.004}
}

@article{xu2026reasoning,
  title={Reasoning-Driven Multimodal {LLM} for Domain 
         Generalization},
  author={Xu, Zhipeng and Wang, Zilong and Jiang, Xinyang and 
          Li, Dongsheng and Cheng, De and Wang, Nannan},
  journal={arXiv preprint arXiv:2602.23777},
  year={2026}
}

@article{gu2021pubmedbert,
  title={Domain-specific language model pretraining for 
         biomedical natural language processing},
  author={Gu, Yu and Tinn, Robert and Cheng, Hao and Lucas, 
          Michael and Usuyama, Naoto and Liu, Xiaodong and 
          Naumann, Tristan and Gao, Jianfeng and Poon, Hoifung},
  journal={ACM Transactions on Computing for Healthcare},
  volume={3},
  number={1},
  pages={1--23},
  year={2022},
  doi={10.1145/3458754}
}

@article{bai2025qwen3vl,
  title={Qwen3-{VL} Technical Report},
  author={Bai, Shuai and Chen, Keqin and Liu, Xuejing and 
          Wang, Jialin and others},
  journal={arXiv preprint arXiv:2511.21631},
  year={2025}
}

@inproceedings{maani2024coreecho,
  title={Coreecho: Continuous representation learning for 2d+ time echocardiography analysis},
  author={Maani, Fadillah Adamsyah and Saeed, Numan and Matsun, Aleksandr and Yaqub, Mohammad},
  booktitle={International Conference on Medical Image Computing and Computer-Assisted Intervention},
  pages={591--601},
  year={2024},
  organization={Springer}
}

@inproceedings{muhtaseb2022echocotr,
  title={Echocotr: Estimation of the left ventricular ejection fraction from spatiotemporal echocardiography},
  author={Muhtaseb, Rand and Yaqub, Mohammad},
  booktitle={International Conference on Medical Image Computing and Computer-Assisted Intervention},
  pages={370--379},
  year={2022},
  organization={Springer}
}

@article{team2024gemma,
  title={Gemma 2: Improving open language models at a practical size},
  author={Team, Gemma and Riviere, Morgane and Pathak, Shreya and Sessa, Pier Giuseppe and Hardin, Cassidy and Bhupatiraju, Surya and Hussenot, L{\'e}onard and Mesnard, Thomas and Shahriari, Bobak and Ram{\'e}, Alexandre and others},
  journal={arXiv preprint arXiv:2408.00118},
  year={2024}
}

@article{abdin2024phi,
  title={Phi-4 technical report},
  author={Abdin, Marah and Aneja, Jyoti and Behl, Harkirat and Bubeck, S{\'e}bastien and Eldan, Ronen and Gunasekar, Suriya and Harrison, Michael and Hewett, Russell J and Javaheripi, Mojan and Kauffmann, Piero and others},
  journal={arXiv preprint arXiv:2412.08905},
  year={2024}
}

\appendix
\renewcommand{\thesection}{\Alph{section}}

\section{Supplementary Material}
\label{sec:supplementary}

\setcounter{table}{0}
\renewcommand{\thetable}{S\arabic{table}}
\setcounter{figure}{0}
\renewcommand{\thefigure}{S\arabic{figure}}

\subsection{Private Dataset}
\label{sec:supp_dataset}
\begin{table*}[t]
\centering
\caption{Private dataset statistics across train and test splits.
\textbf{Left:} abnormality label and view distributions.
\textbf{Right:} QA pair distribution and detector annotation
statistics before and after RAFT propagation~\cite{teed2020raft}.
$^\dagger$GRPO uses only Abnormality List and Full Report question types.}
\label{tab:dataset_full}
\setlength{\tabcolsep}{5pt}
\renewcommand{\arraystretch}{1.05}

\begin{minipage}[t]{0.47\textwidth}
\centering
\resizebox{\textwidth}{!}{%
\begin{tabular}{l rr rr}
\toprule
& \multicolumn{2}{c}{\textbf{Train} ($n\!=\!5{,}061$)}
& \multicolumn{2}{c}{\textbf{Test} ($n\!=\!1{,}215$)} \\
\cmidrule(lr){2-3} \cmidrule(lr){4-5}
& Pos. & Neg. & Pos. & Neg. \\
\midrule
\multicolumn{5}{l}{\cellcolor{gray!10}\textbf{Abnormality Labels}} \\
TV Regurgitation        & 3,291 & 1,770 &  662 &  553 \\
MV Regurgitation        & 3,133 & 1,928 &  659 &  556 \\
AV Regurgitation        & 1,191 & 3,870 &  188 & 1,027 \\
LA Enlargement          & 1,694 & 3,367 &  330 &  885 \\
LV Systolic Dysfunction & 1,070 & 3,991 &  198 & 1,017 \\
MV Calcification        &   701 & 4,360 &  105 & 1,110 \\
LV Enlargement          &   504 & 4,557 &   70 & 1,145 \\
AV Stenosis             &   481 & 4,580 &  116 & 1,099 \\
RA Enlargement          &   384 & 4,677 &   59 & 1,156 \\
Bicuspid AV             &   230 & 4,831 &   31 & 1,184 \\
RV Enlargement          &   221 & 4,840 &   31 & 1,184 \\
Normal                  &   556 &    -- &  212 &    -- \\
\midrule
\multicolumn{5}{l}{\cellcolor{gray!10}\textbf{View Distribution}} \\
& Clips & \% & Clips & \% \\
\cmidrule(lr){2-3} \cmidrule(lr){4-5}
A2C                     & 5,633 & 12.1 & 1,409 & 12.5 \\
A2C Color on MV         & 2,997 &  6.5 &   693 &  6.1 \\
A3C                     & 6,069 & 13.1 & 1,485 & 13.1 \\
A3C Color on AV         & 2,351 &  5.1 &   548 &  4.8 \\
A4C                     & 7,087 & 15.3 & 1,911 & 16.9 \\
A4C Zoomed LV           & 3,374 &  7.3 &   750 &  6.6 \\
PLAX Standard           & 7,485 & 16.1 & 1,770 & 15.6 \\
PLAX Mitral Cusps       & 3,158 &  6.8 &   780 &  6.9 \\
PSAX Zoomed Out         & 5,665 & 12.2 & 1,316 & 11.6 \\
PSAX Color on Pulm.     & 2,563 &  5.5 &   648 &  5.7 \\
\rowcolor{gray!6}
\textbf{Total}          & 46,382 & 100 & 11,310 & 100 \\
\bottomrule
\end{tabular}}
\end{minipage}
\hfill
\begin{minipage}[t]{0.51\textwidth}
\centering
\resizebox{\textwidth}{!}{%
\begin{tabular}{l rr rr}
\toprule
& \multicolumn{2}{c}{\textbf{Train}}
& \multicolumn{2}{c}{\textbf{Test}} \\
\cmidrule(lr){2-3} \cmidrule(lr){4-5}
\textbf{Question Type} & Pairs & \% & Pairs & \% \\
\midrule
\multicolumn{5}{l}{\cellcolor{gray!10}\textbf{QA Pairs}} \\
Structure Description      & 57,622 & 44.9 & 14,230 & 45.6 \\
Abnormality Classification & 55,671 & 43.4 & 13,365 & 42.8 \\
Abnormality List           &  5,061 &  3.9 &  1,215 &  3.9 \\
Conclusion                 &  5,061 &  3.9 &  1,215 &  3.9 \\
Full Report      &  4,800 &  3.7 &  1,184 &  3.8 \\
\rowcolor{gray!6}
\textbf{Total}             & 128,215 & 100 & 31,209 & 100 \\
\midrule
\multicolumn{5}{l}{\cellcolor{gray!10}\textbf{Detector Annotations, n frames}} \\
& Pre-RAFT && \multicolumn{2}{c}{Post-RAFT} \\
\cmidrule(lr){2-2} \cmidrule(lr){4-5}
& Annotated && Train & Test \\
\cmidrule(lr){2-2} \cmidrule(lr){4-5}
Left Ventricle (LV)         &  35,365 && 319,375 &  50,898 \\
Left Atrium  (LA)           &   7,907 &&  86,207 &  13,014 \\
Right Atrium  (RA)          &   3,958 &&  36,290 &   5,470 \\
Right Ventricle (RV)        &   3,018 &&  33,011 &   5,038 \\
Mitral Valve  (MV)          &   1,309 &&  10,312 &   1,367 \\
Tricuspid Valve  (TV)       &   1,294 &&  10,007 &   1,229 \\
LVOT                    &     519 &&   7,540 &     983 \\
\rowcolor{gray!6}
\textbf{Total}          &  53,370 && 502,742 &  77,999 \\
\midrule
\multicolumn{5}{l}{\cellcolor{gray!10}\textbf{Coverage Statistics}} \\
& Pre-RAFT && Train & Test \\
\cmidrule(lr){2-2} \cmidrule(lr){4-5}
DICOMs                  &  10,291 &&   7,964 &   1,245 \\
Total frames            &  384,456 && 332,047 &  52,674 \\
Avg.\ structs / frame   &     1.5 &&    1.52 &    1.48 \\
\bottomrule
\end{tabular}}
\end{minipage}
\end{table*}

The private echocardiography dataset was collected between 2018 and 2022 from
139 sites and comprises 5,061 training and 1,215 test studies, containing
46,382 and 11,310 videos, respectively. Each study consists of multi-view
echocardiographic clips acquired across up to 10 standard views. At training
and inference time, one clip per unique view is randomly selected per study,
providing implicit data augmentation across epochs. Per-study clip counts range
from 4 to 30 in training and 4 to 22 at test time, with a median of 9 clips
per study. Full dataset statistics are reported in ~\cref{tab:dataset_full}.

\paragraph{Abnormality Labels.}
Each study is annotated with binary labels for 12 cardiac abnormalities spanning
valvular disease, chamber enlargement, systolic dysfunction, and structural
anomalies. TV and MV Regurgitation are the most prevalent, while Bicuspid AV and RV Enlargement are the rarest. 

\paragraph{View Distribution.}
The dataset spans 10 standard echocardiographic views: seven B-mode views
(PLAX Standard, A4C, A3C, PSAX Zoomed Out, A2C, A4C Zoomed LV, PLAX Mitral
Cusps) and three color Doppler variants (A2C Color on MV, PSAX Color on
Pulmonary, A3C Color on AV). The train and test splits follow closely matched
distributions across all views.

\begin{table*}[t]
\centering
\caption{Representative question--answer examples per type used in SFT and GRPO
training. $^\dagger$Used in GRPO only. All examples are drawn from the private
training set.}
\label{tab:question_examples}
\setlength{\tabcolsep}{5pt}
\renewcommand{\arraystretch}{1.05}
\resizebox{\textwidth}{!}{%
\begin{tabular}{l p{5.5cm} p{11cm}}
\toprule
\textbf{Type} & \textbf{Example Question} & \textbf{Example Answer} \\
\midrule

\textbf{Structure Description}
  & What are the echocardiographic findings for the left atrium?
  & \textit{Normal dimensions. No suggestion of mass or thrombus.} \\[3pt]
  & Describe the tricuspid valve findings in this echocardiographic study.
  & \textit{Normal appearance. Mild tricuspid regurgitation.} \\[3pt]
  & What does the echocardiogram reveal about the aortic root?
  & \textit{Normal dimensions for BSA.} \\
\midrule

\textbf{Abnormality Classification}
  & Is there tricuspid regurgitation present in this echocardiogram?
  & \textit{Yes, tricuspid regurgitation is present.} \\[3pt]
  & Is the left atrium enlarged?
  & \textit{No, there is no evidence of left atrial enlargement.} \\[3pt]
  & Is there left ventricular systolic dysfunction present in this echocardiogram?
  & \textit{No, there is no evidence of left ventricular systolic dysfunction.} \\
\midrule

\textbf{Abnormality List}
  & Summarize the abnormalities observed in this echocardiographic examination.
  & \textit{The following abnormalities are identified: Tricuspid Regurgitation, Mitral Valve Regurgitation.} \\[3pt]
  & What abnormalities are present in this echocardiographic study?
  & \textit{No significant abnormalities identified.} \\
\midrule

\textbf{Conclusion}
  & What is the overall conclusion of this echocardiographic study?
  & \textit{Normal left ventricular size and function. Mild mitral regurgitation. Stage I diastolic dysfunction with normal LV filling pressure.} \\[3pt]
  & Summarize the key conclusions from this echocardiographic examination.
  & \textit{No evidence of hypertrophy, chamber enlargement, or significant valvular disease.} \\
\midrule

\textbf{Full Report}
  & Produce a detailed echocardiographic report with findings for each structure and a final conclusion.
  & \textit{\begin{tabular}[t]{@{}>{\raggedright\arraybackslash}p{11cm}@{}}
    Left Ventricle: Normal dimensions and systolic function. Normal wall thickness. Impaired LV relaxation consistent with Stage I diastolic dysfunction. LV filling pressure is normal.\\
    Right Ventricle: Normal dimensions and systolic function.\\
    Left Atrium: Normal dimensions. No suggestion of mass or thrombus.\\
    Right Atrium: Normal dimensions. No suggestion of mass or thrombus.\\
    Aortic Valve: Normal trileaflet morphology. Normal Doppler flow characteristics.\\
    Mitral Valve: Normal appearance without prolapse. Mild mitral regurgitation.\\
    Tricuspid Valve: Normal appearance. Normal Doppler flow characteristics.\\
    Pulmonic Valve: Normal appearance. Normal Doppler flow characteristics.\\
    Pericardium: Appears normal. No significant effusion is seen.\\
    Aortic Root: Normal dimensions for BSA.\\
    Aortic Arch: Normal dimensions. Normal Doppler flow profile.\\
    Pulmonary Artery: RVSP could not be estimated because of insufficient tricuspid regurgitation.\\
    Conclusions: Normal left ventricular size with estimated ejection fraction of 60\%. No evidence of hypertrophy, chamber enlargement or aortic dilatation. Mild mitral regurgitation. RVSP could not be estimated because of insufficient tricuspid regurgitation. Stage I diastolic dysfunction with normal LV filling pressure. Normal peak velocities noted through the aortic valve, left ventricular outflow tract, pulmonic valve and descending aorta.
  \end{tabular}} \\
\bottomrule
\end{tabular}}
\end{table*}
\paragraph{Training Data.}
For SFT, the training set yields 128,215 QA pairs across 5,061 studies (mean
25.3 pairs per study), spanning all five question types. For GRPO, only the
abnormality list and full report question types are used, as these directly
optimize for the two target tasks. ~\cref{tab:question_examples}
shows representative examples of each question type.

\paragraph{Structure-Aware Detector Annotations.}
A subset of 2,689 training studies was manually annotated with segmentation masks for seven cardiac
structures: LV, LA, RV, RA, MV, TV, and LVOT. Masks were converted to bounding
boxes and propagated across all frames using RAFT optical
flow, expanding from sparsely annotated keyframes to full
per-frame coverage.

\subsection{Reasoning Trace Generation}
\label{sec:supp_reasoning}

Reasoning traces for all training samples are generated offline
with Qwen3-8B in a text-only setting, with temperature~$= 0.6$, top-$p = 0.95$. The model is given the correct answer upfront and asked to reason backward toward it.

\paragraph{Prompt Construction.}
For each study, a structured prompt is constructed containing: (i) a system
message identifying the model as an expert echocardiographer; (ii) the
ground-truth disease label or list of present abnormalities; (iii) the list of
available echocardiographic views; and for each view: (iv) detected cardiac
structures, (v) pathology-specific assessment guidance, and (vi) clinical
findings from the ground-truth report. The prompt concludes with an instruction
to analyze each view and synthesize a final assessment.

The assessment guidance is encoded as per-pathology diagnostic templates that
specify, for each view, its diagnostic importance and the structures and features
that are clinically relevant for a given condition. For example, the Aortic
Stenosis template designates PLAX Standard for leaflet morphology and
calcification assessment, PSAX Zoomed Out for valve orifice planimetry and cusp
count, and A3C Color on AV for peak velocity and mean gradient via CW Doppler.
Templates are defined for all 11 target abnormalities, 12 cardiac structures,
and a comprehensive all-pathology view set, covering both B-mode and color
Doppler views, and are derived from clinical
guidelines.

\paragraph{Reasoning Trace Construction.}
The generated analyses are mapped to VQA training samples as follows.
For \emph{abnormality classification} questions, the disease-specific
analysis for the queried condition is used. For \emph{structure
description} questions, the structure-specific section of the analysis
is used. For \emph{abnormality list}, \emph{conclusion}, and \emph{full
report} questions, the analysis is generated using a view-complete
template covering all cardiac structures across all views simultaneously.
Each reasoning trace is wrapped in
\texttt{<think>}$\ldots$\texttt{</think>} tags and prepended to the
ground-truth answer, forming the complete supervised target
$\mathcal{Y}^*$.

\paragraph{Reasoning Trace Quality Audit.}
To validate the quality of the generated training traces, we audit 1,000 randomly sampled traces using two independent judges (Gemma-2-9B-it~\cite{team2024gemma} and Phi-4-14B~\cite{abdin2024phi}) across six dimensions: fabrication, coverage, visual grounding, view-pathology appropriateness, internal consistency, and terminology accuracy. Both judges score visual grounding and view-pathology appropriateness below terminology accuracy (Gemma: 4.885, 4.860 vs.\ 4.998; Phi-4: 4.872, 4.942 vs.\ 4.980), indicating the audit is not at ceiling; all dimensions nonetheless score above 4.73/5. These results confirm that the LLM structures clinical content already present in the prompt rather than fabricating new findings, as the traces are anchored on per-pathology assessment templates derived from American Society of Echocardiography guidelines~\cite{lang2015ase}, ground-truth report findings, detector-confirmed structures, and the available view list.

\subsection{Structure-Aware Detector}
\label{sec:supp_detector}

\paragraph{Architecture and Training Data.}
The structure-aware detector is an RT-DETR-L model fine-tuned from
COCO-pretrained weights to localize seven cardiac structures: LV, LA,
RA, RV, MV, TV, and LVOT. Training uses RAFT-propagated bounding boxes
(~\cref{sec:supp_dataset}) at $640\times640$ resolution. Given
the substantial annotation imbalance between common structures (LV, LA)
and rarer ones (RA, RV, MV, TV, LVOT), we apply class-balanced sampling:
all frames containing rare structures are retained, with an equal number
of LV/LA-only frames randomly sampled to match.

\paragraph{Training Configuration.}
The detector is trained for 5 epochs with batch
size 64, AdamW optimizer (lr$_0 = 0.01$, lr$_f = 0.01$), and AMP enabled.
Augmentation includes mosaic ($p=1.0$), horizontal flip ($p=0.5$), random scale
($\pm 0.5$), random translate ($\pm 0.1$), random erasing ($p=0.4$), HSV
jitter, and RandAugment. IoU threshold is set to 0.7.
Best validation performance is achieved at epoch 2, with mAP$_{50} = 0.350$
and mAP$_{50\text{-}95} = 0.161$ (precision $0.304$, recall $0.619$).

\begin{figure*}[t]
    \centering
    \includegraphics[width=0.85\textwidth]{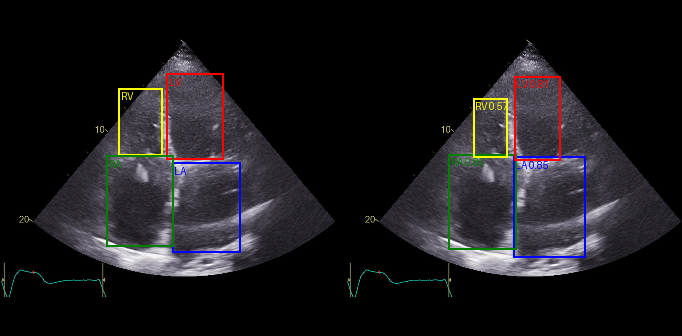}
    
    \vspace{4pt}
    
    \includegraphics[width=0.85\textwidth]{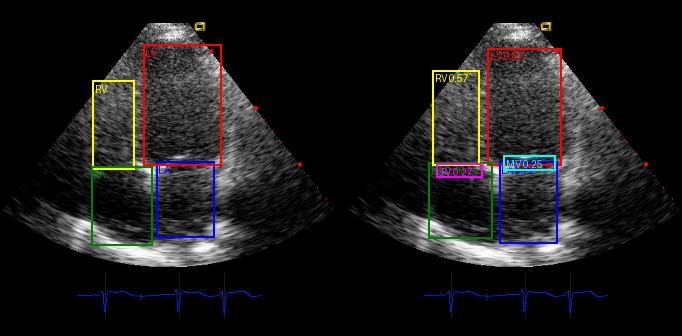}
    \caption{Qualitative detection results on the private dataset. In each
    pair, the \textbf{left} image shows ground-truth bounding boxes and the
    \textbf{right} shows model predictions with confidence scores. Colour
    coding: \textcolor{red}{$\blacksquare$}~LV,
    \textcolor{green}{$\blacksquare$}~LA,
    \textcolor{blue}{$\blacksquare$}~RA,
    \textcolor{yellow}{$\blacksquare$}~RV,
    \textcolor{magenta}{$\blacksquare$}~MV,
    \textcolor{cyan}{$\blacksquare$}~TV,
    \textcolor{orange}{$\blacksquare$}~LVOT.}
    \label{fig:detector_private}
\end{figure*}

\begin{figure*}[t]
    \centering
    \begin{tabular}{cc}
        \includegraphics[width=0.42\textwidth]{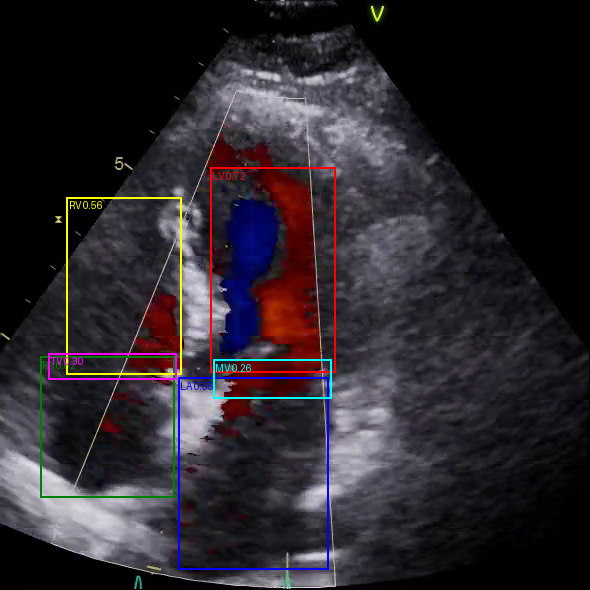} &
        \includegraphics[width=0.42\textwidth]{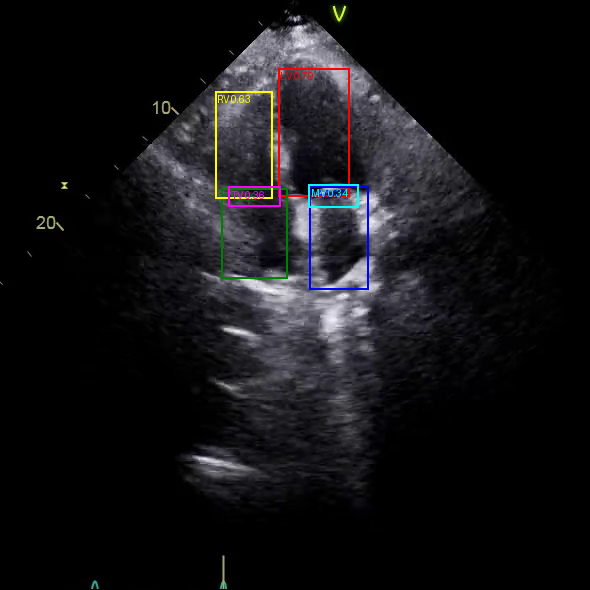} \\[4pt]
        \multicolumn{2}{c}{\small\textit{MIMICEchoQA}} \\[8pt]
        \includegraphics[width=0.42\textwidth]{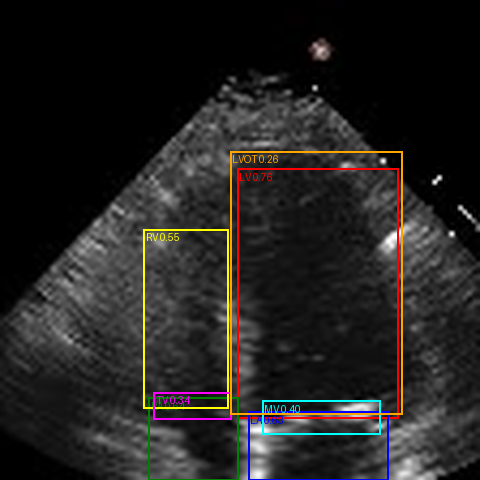} &
        \includegraphics[width=0.42\textwidth]{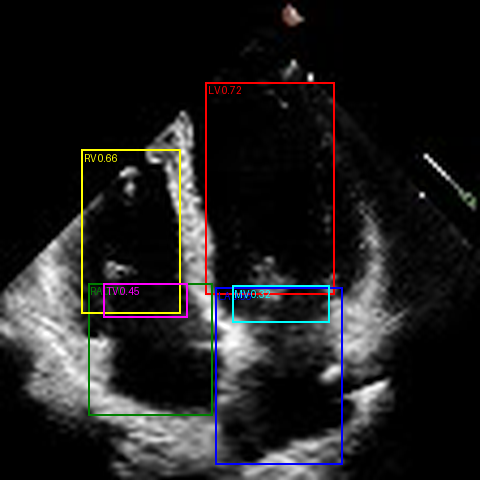} \\[4pt]
        \multicolumn{2}{c}{\small\textit{EchoNet-Dynamic}} \\
    \end{tabular}
    \caption{Zero-shot generalisation of the structure-aware detector to
    public benchmarks unseen during training. Colour coding as in
    ~\cref{fig:detector_private}.}
    \label{fig:detector_public}
\end{figure*}

\paragraph{Qualitative Results.}
~\cref{fig:detector_private} shows representative detection results
on the private dataset. The model correctly localises the dominant
structures (LV, RV, LA) across views, with predicted boxes closely
matching ground truth. Smaller structures such as MV and TV are
detected with lower confidence but remain anatomically consistent.
~\cref{fig:detector_public} demonstrates generalisation
to MIMICEchoQA and EchoNet-Dynamic, both unseen during training.
Despite differences in image quality, scanner settings, and the
presence of color Doppler overlays, the detector produces anatomically
plausible localisations across all four examples, suggesting that the
learned structural representations transfer across institutions and
acquisition protocols.

\subsection{View Classifier}
\label{sec:supp_viewclassifier}

The view classifier assigns a standard echocardiographic view label to each
incoming video, providing the textual view identifiers interleaved with
visual tokens in the multi-view input sequence (\cref{sec:encoding}).
It is based on MobileNet-V3-Small, fine-tuned from ImageNet-pretrained weights on the private training set using a single random frame per video per epoch. Training uses AdamW with differential learning rates (backbone $10^{-4}$, head $10^{-3}$) and cosine annealing over 15 epochs.
On the private test set, the classifier
achieves 95.7\% accuracy and 95.7\% macro F1, with well-calibrated
confidence (ECE $= 0.0078$). Per-view F1 ranges from 0.888 (A4C Zoomed LV)
to 0.992 (PSAX Color on Pulmonary). The most frequent confusions occur between
visually similar views: A4C and A4C Zoomed LV,
A2C and A4C, which share similar apical geometry and are
distinguishable primarily by field of view.

\subsection{Evaluation Prompts}
\label{sec:supp_prompts}

At evaluation time, all models are prompted to perform two tasks:
abnormality classification and report generation. Prompts differ between
EchoSonar-R and the baseline VLMs to accommodate their respective input
formats and output conventions.

\paragraph{Abnormality Classification.}
EchoSonar-R and its ablations are prompted with an open-ended question,
consistent with the SFT training format:

\begin{tcolorbox}[promptbox]
What abnormalities are present in this echocardiographic study?
\end{tcolorbox}

Baseline VLMs receive a structured prompt to target
abnormalities explicitly, as these models have not been trained to produce
a free-form abnormality list:

\begin{tcolorbox}[promptbox]
Based on this echocardiographic study, identify which of the following
abnormalities are present or absent: Tricuspid Regurgitation, Mitral Valve Regurgitation, Aortic Stenosis, Mitral Valve Calcification, Left Atrial Enlargement, Left Ventricular Enlargement, Left Ventricular Systolic Dysfunction, Right Atrial Enlargement, Right Ventricular Enlargement, Bicuspid Aortic Valve, Aortic Regurgitation
For each abnormality, state whether it is Present or Absent.
\end{tcolorbox}

\paragraph{Report Generation.}
For report generation, the following prompt is used:

\begin{tcolorbox}[promptbox]
You are an expert echocardiographer. Based on the provided
echocardiographic images, generate a structured echocardiographic
report with the following sections. For each section, provide a
concise clinical description of the findings.\\[4pt]
Left Ventricle: Right Ventricle:
Left Atrium: Right Atrium:
Aortic Valve: Mitral Valve:
Tricuspid Valve: Pulmonic Valve:
Pericardium: Aortic Root:
Aortic Arch: Pulmonary Artery:
Conclusions:\\[4pt]
Write each section as a brief clinical narrative. If a structure is
not well visualized, state that.
\end{tcolorbox} 
\noindent Predicted abnormality lists and report sections are extracted
from model outputs via string matching and section parsing, respectively.
For EchoSonar-R, only the content after \texttt{</think>} is used for
evaluation; the reasoning trace is excluded.

\paragraph{Reasoning Quality Evaluation.}
Reasoning trace quality is assessed using an LLM-as-a-judge framework on five clinical dimensions,
each rated on a 1--5 Likert scale. The evaluation prompt
provides the original question, ground-truth answer, and full model
response. The five dimensions are:

\begin{enumerate}[leftmargin=*, itemsep=2pt]
    \item Reasoning Efficiency: whether every reasoning step
    advances toward the answer, penalising redundancy and circular
    reasoning.
    \item Factual Correctness: accuracy of clinical and
    anatomical claims made during reasoning.
    \item Evidence Grounding: whether the model references
    specific visual observations rather than relying on generic
    statements.
    \item Terminology Accuracy: correct use of
    echocardiographic and medical terminology throughout.
    \item Reasoning-Answer Agreement: consistency between
    findings discussed in the reasoning trace and those stated in the
    final answer, penalising both omissions and unsupported additions.
\end{enumerate}

\begin{tcolorbox}[promptbox, title=Reasoning Quality Prompt,
    colbacktitle=gray!40, coltitle=black,
    label=box:reasoning_prompt,
    float, floatplacement=t,
    fontupper=\footnotesize]
You are an expert echocardiographer evaluating the quality of clinical
reasoning in echocardiographic analysis.\\[4pt]
Task given to the model: \{question\}\\
Ground truth answer: \{ground\_truth\}\\
Model's response: \{response\}\\[4pt]
Evaluate the response on these 5 dimensions (score 1-5 each):\\[4pt]
{[Reasoning Efficiency]}\\
\hspace*{1em}5: Concise and focused - every step advances toward the answer\\
\hspace*{1em}4: Mostly focused with minor digressions\\
\hspace*{1em}3: Some unnecessary repetition or tangents\\
\hspace*{1em}2: Significant redundancy or circular reasoning\\
\hspace*{1em}1: Rambling, mostly irrelevant\\[3pt]
{[Factual Correctness]}\\
\hspace*{1em}5: All clinical/anatomical facts are correct\\
\hspace*{1em}4: One minor factual inaccuracy\\
\hspace*{1em}3: A few factual errors that don't affect the conclusion\\
\hspace*{1em}2: Multiple factual errors, some affecting the conclusion\\
\hspace*{1em}1: Pervasive factual errors\\[3pt]
{[Evidence Grounding]}\\
\hspace*{1em}5: Consistently references specific observations (chamber sizes, wall\\
\hspace*{2em}motion, valve appearance, Doppler patterns)\\
\hspace*{1em}4: References some specific observations\\
\hspace*{1em}3: Mix of specific and generic statements\\
\hspace*{1em}2: Mostly generic reasoning with few image references\\
\hspace*{1em}1: No reference to actual image content\\[3pt]
{[Terminology Accuracy]}\\
\hspace*{1em}5: All medical/echocardiographic terms used correctly\\
\hspace*{1em}4: One minor terminological imprecision\\
\hspace*{1em}3: A few terminology issues that don't cause confusion\\
\hspace*{1em}2: Multiple misused terms (e.g., stenosis vs.\ regurgitation)\\
\hspace*{1em}1: Pervasive terminology errors\\[3pt]
{[Reasoning-Answer Agreement]}\\
\hspace*{1em}5: Perfect agreement - every finding in reasoning reflected in answer\\
\hspace*{1em}4: Minor gap - one finding missing from answer or lacking reasoning\\
\hspace*{1em}3: Partial agreement - some findings dropped or answer includes\\
\hspace*{2em}items not discussed in reasoning\\
\hspace*{1em}2: Poor agreement - reasoning and answer substantially diverge\\
\hspace*{1em}1: No agreement - reasoning and answer contradict each other\\[4pt]
Respond in this format:\\
{[Reasoning Efficiency]} \textlangle score\textrangle\\
{[Factual Correctness]} \textlangle score\textrangle\\
{[Evidence Grounding]} \textlangle score\textrangle\\
{[Terminology Accuracy]} \textlangle score\textrangle\\
{[Reasoning-Answer Agreement]} \textlangle score\textrangle\\
{[Explanation]} \textlangle brief explanation\textrangle
\end{tcolorbox}

\paragraph{Clinical Report Faithfulness.}
Report generation quality is assessed using an echocardiography-adapted
version of the GREEN metric~\cite{ostmeier2024green}, evaluated
section-by-section. For each
predicted report section, the judge identifies clinically significant
errors (false findings, omissions, wrong anatomical location,
misassessed severity, incorrect comparison to prior studies) and
clinically insignificant errors (omission of minor findings, wording
differences that do not change clinical meaning), and counts matched
findings. The GREEN score is then computed from these counts as described
in~\cite{ostmeier2024green}. This approach is more sensitive to clinical
correctness than lexical metrics such as BLEU or ROUGE, which penalise
semantically equivalent phrasings. The full evaluation prompt is shown
below.

\begin{tcolorbox}[promptbox, title=Echo-GREEN Prompt,
    colbacktitle=gray!40, coltitle=black,
    label=box:green_prompt,
    float, floatplacement=t,
    fontupper=\footnotesize]
You are an expert echocardiographer evaluating the accuracy of a
candidate echocardiographic report section against a reference.\\[4pt]
Your task:\\
\hspace*{1em}1. Identify all clinically significant errors in the candidate.\\
\hspace*{1em}2. Identify all clinically insignificant errors in the candidate.\\
\hspace*{1em}3. Count the number of matched findings (correct statements).\\[4pt]
{[Clinically Significant Errors]}\\
\hspace*{1em}(a) False finding in candidate not present in reference.\\
\hspace*{1em}(b) Missing finding present in reference but absent from candidate.\\
\hspace*{1em}(c) Wrong anatomical location (e.g., wrong valve or chamber).\\
\hspace*{1em}(d) Misassessed severity or function (e.g., mild vs.\ severe).\\
\hspace*{1em}(e) False comparison to a prior study not in the reference.\\
\hspace*{1em}(f) Omitted comparison to a prior study mentioned in the reference.\\[3pt]
{[Clinically Insignificant Errors]}\\
\hspace*{1em}(a) Omission of a clinically insignificant finding.\\
\hspace*{1em}(b) Inclusion of a clinically insignificant finding not in reference.\\
\hspace*{1em}(c) Minor wording differences that do not change clinical meaning.\\[4pt]
Respond in this format:\\
{[Clinically Significant Errors]}\\
\hspace*{1em}(a) \textlangle count\textrangle\ $\cdots$ (f) \textlangle count\textrangle\\
\hspace*{1em}Total: \textlangle total\textrangle\\
{[Clinically Insignificant Errors]}\\
\hspace*{1em}(a) \textlangle count\textrangle\ $\cdots$ (c) \textlangle count\textrangle\\
\hspace*{1em}Total: \textlangle total\textrangle\\
{[Matched Findings]} \textlangle count\textrangle\\
{[Explanation]} \textlangle brief explanation\textrangle\\[4pt]
Reference: \{reference\}\\
Candidate: \{candidate\}
\end{tcolorbox}

\subsection{Additional Results}
\label{sec:supp_results}
\paragraph{Per-Abnormality Results.}
~\cref{tab:classification_extended} reports per-abnormality F1,
balanced accuracy (BAcc), sensitivity, and specificity across all three
evaluation sets. On the private test set, EchoSonar-R achieves the
highest F1 on 9 of 12 conditions, with the largest margins on rare or
structurally complex abnormalities such as Aortic Stenosis, Bicuspid AV,
and LV Enlargement. The high specificity across these rare conditions
(98-99\%) confirms that the model avoids over-predicting low-prevalence
pathologies. On MIMICEchoQA, EchoSonar-R leads on LV Systolic
Dysfunction and Aortic Regurgitation, though performance on Right
Ventricular Enlargement and Left Atrial Enlargement is lower, consistent
with the label shift discussed in the main paper. On EchoNet-Dynamic,
which evaluates only EF-based LV systolic dysfunction from a single
apical view, EchoSonar-R achieves the highest BAcc (60.2\%).

\begin{table*}[t]
\centering
\caption{Abnormality classification performance across private and
public datasets. We report per-disease F1 score ($\uparrow$),
balanced accuracy (BAcc, $\uparrow$), sensitivity (Se, $\uparrow$),
and specificity (Sp, $\uparrow$).
EchoSonar-R$^{\dagger}$ denotes GRPO training;
EchoSonar-R$^{\ast}$ denotes SFT-only.
Baselines: Qwen3-VL~\cite{bai2025qwen3vl},
MedGemma~\cite{sellergren2025medgemma},
EchoVLM~\cite{she2025echovlm},
Chiron-o1~\cite{sun2025chirono1},
Lingshu~\cite{xu2025lingshu}.
Prev.\ = prevalence (\%).}
\label{tab:classification_extended}
\setlength{\tabcolsep}{3.5pt}
\resizebox{\textwidth}{!}{%
\begin{tabular}{l c | cccc | cccc || cccc | cccc | cccc | cccc | cccc}
\toprule
& & \multicolumn{4}{c|}{\textbf{EchoSonar-R}$^{\dagger}$}
  & \multicolumn{4}{c||}{\textbf{EchoSonar-R}$^{\ast}$}
  & \multicolumn{4}{c|}{\textbf{Qwen3-VL}}
  & \multicolumn{4}{c|}{\textbf{MedGemma}}
  & \multicolumn{4}{c|}{\textbf{EchoVLM}}
  & \multicolumn{4}{c|}{\textbf{Chiron-o1}}
  & \multicolumn{4}{c}{\textbf{Lingshu}} \\
\cmidrule(lr){3-6}\cmidrule(lr){7-10}\cmidrule(lr){11-14}
\cmidrule(lr){15-18}\cmidrule(lr){19-22}\cmidrule(lr){23-26}\cmidrule(lr){27-30}
\textbf{Abnormality} & \textbf{Prev.}
  & F1 & BAcc & Se & Sp
  & F1 & BAcc & Se & Sp
  & F1 & BAcc & Se & Sp
  & F1 & BAcc & Se & Sp
  & F1 & BAcc & Se & Sp
  & F1 & BAcc & Se & Sp
  & F1 & BAcc & Se & Sp \\
\midrule
\multicolumn{30}{l}{\cellcolor{yellow!8}\textit{\textbf{Private Test Set} ($n = 1{,}215$ studies, multi-view)}} \\[3pt]
TV Regurgitation         & 54.6 & 66.6 & 59.9 & 71.5 & 48.3 & 66.6 & 59.3 & 72.4 & 46.3 & 38.3 & 52.1 & 28.5 & 75.6 & 27.9 & 51.2 & 18.4 & 83.9 & 0.0  & 50.0 & 0.0  & 100.0& 61.1 & 48.8 & 70.8 & 26.8 & 62.3 & 49.4 & 73.4 & 25.3 \\[1.5pt]
MV Regurgitation         & 54.4 & 68.3 & 61.6 & 74.3 & 48.9 & 69.1 & 62.6 & 75.1 & 50.0 & 49.7 & 51.5 & 44.6 & 58.5 & 24.6 & 49.2 & 16.1 & 82.4 & 0.0  & 50.0 & 0.0  & 100.0& 69.6 & 50.5 & 96.1 & 5.0  & 33.6 & 50.5 & 24.1 & 76.8 \\[1.5pt]
LA Enlargement           & 27.3 & 64.9 & 75.8 & 64.8 & 86.9 & 62.7 & 74.6 & 65.5 & 83.8 & 26.6 & 49.8 & 26.4 & 73.3 & 35.1 & 49.3 & 51.2 & 47.5 & 0.0  & 50.0 & 0.0  & 100.0& 40.2 & 49.5 & 79.7 & 19.2 & 29.5 & 48.0 & 35.8 & 60.3 \\[1.5pt]
Healthy                  & 17.5 & 38.0 & 62.3 & 35.2 & 89.4 & 36.2 & 61.3 & 33.0 & 89.5 & 20.1 & 49.9 & 24.1 & 75.7 & 23.7 & 48.1 & 43.9 & 52.2 & 0.0  & 50.0 & 0.0  & 100.0& 0.0  & 49.9 & 0.0  & 99.8 & 1.9  & 50.4 & 0.9  & 99.8 \\[1.5pt]
LV Systolic Dysfunction  & 16.3 & 59.5 & 74.0 & 52.8 & 95.2 & 55.9 & 71.9 & 49.0 & 94.9 & 21.8 & 51.8 & 26.8 & 76.8 & 19.6 & 48.6 & 28.3 & 68.8 & 0.0  & 50.0 & 0.0  & 100.0& 27.9 & 50.5 & 92.4 & 8.6  & 28.0 & 51.2 & 86.4 & 16.1 \\[1.5pt]
AV Regurgitation         & 15.4 & 42.3 & 65.4 & 39.4 & 91.5 & 41.3 & 65.1 & 39.9 & 90.3 & 24.2 & 52.3 & 41.5 & 63.2 & 16.8 & 50.4 & 17.6 & 83.2 & 26.8 & 50.0 &100.0 & 0.0  & 26.1 & 50.0 & 82.4 & 17.5 & 25.6 & 49.9 & 75.0 & 24.7 \\[1.5pt]
AV Stenosis              &  9.5 & 71.6 & 82.2 & 66.4 & 98.0 & 69.8 & 80.9 & 63.8 & 98.0 & 14.9 & 50.3 & 31.9 & 68.7 & 10.0 & 49.7 & 11.2 & 88.2 & 5.8  & 50.9 & 3.4  & 98.3 & 16.6 & 50.9 & 51.7 & 50.1 & 18.5 & 54.7 & 30.2 & 79.3 \\[1.5pt]
MV Calcification         &  8.6 & 45.4 & 72.2 & 51.4 & 92.9 & 44.5 & 69.7 & 44.8 & 94.7 & 15.4 & 51.7 & 45.7 & 57.7 & 9.4  & 48.1 & 14.3 & 82.0 & 0.0  & 50.0 & 0.0  & 100.0& 14.1 & 47.7 & 57.1 & 38.2 & 15.9 & 51.2 & 76.2 & 26.2 \\[1.5pt]
LV Enlargement           &  5.7 & 57.4 & 75.5 & 52.9 & 98.1 & 53.9 & 71.5 & 44.3 & 98.8 & 11.2 & 52.5 & 42.9 & 62.1 & 8.8  & 47.9 & 31.4 & 64.5 & 0.0  & 50.0 & 0.0  & 100.0& 11.1 & 51.1 & 92.9 & 9.4  & 5.0  & 46.2 & 10.0 & 82.4 \\[1.5pt]
RA Enlargement           &  4.8 & 25.9 & 58.7 & 18.6 & 98.7 & 18.4 & 55.9 & 13.6 & 98.3 & 6.0  & 46.9 & 16.9 & 76.9 & 9.1  & 50.9 & 50.8 & 50.9 & 0.0  & 50.0 & 0.0  & 100.0& 8.9  & 48.4 & 88.1 & 8.7  & 8.7  & 48.0 & 76.3 & 19.6 \\[1.5pt]
RV Enlargement           &  2.5 & 13.8 & 55.5 & 12.9 & 98.1 & 3.6  & 50.6 & 3.2  & 98.0 & 2.5  & 44.2 & 12.9 & 75.6 & 2.3  & 41.1 & 16.1 & 66.1 & 0.0  & 50.0 & 0.0  & 100.0& 5.4  & 53.8 &100.0 & 7.7  & 5.2  & 52.1 & 83.9 & 20.3 \\[1.5pt]
Bicuspid AV              &  2.5 & 39.2 & 65.7 & 32.3 & 99.2 & 19.2 & 57.4 & 16.1 & 98.6 & 4.5  & 50.1 & 19.4 & 80.8 & 6.7  & 55.6 & 35.5 & 75.8 & 0.0  & 50.0 & 0.0  & 100.0& 5.2  & 51.9 & 67.7 & 36.1 & 5.1  & 51.4 & 67.7 & 35.0 \\[1.5pt]
\rowcolor{yellow!12}
Macro Average            &      & 49.4 & 67.4 & 47.7 & 87.1 & 45.1 & 65.1 & 43.4 & 86.8 & 19.6 & 50.3 & 30.1 & 70.4 & 16.2 & 49.2 & 27.9 & 70.5 & 2.7  & 50.1 & 8.6  & 91.5 & 23.9 & 50.3 & 73.2 & 27.3 & 19.9 & 50.3 & 53.3 & 47.2 \\
\midrule
\multicolumn{30}{l}{\cellcolor{yellow!8}\textit{\textbf{MIMICEchoQA}~\cite{thapa2025mimicechoqa} (single-view)}} \\[3pt]
LA Enlargement, $n=18$        & 77.8 & 11.6 & 52.7 & 28.6 & 76.9 & 6.1  & 45.8 & 14.3 & 77.4 & 58.3 & 37.5 & 50.0 & 25.0 & 66.7 & 32.1 & 64.3 & 0.0  & 0.0  & 50.0 & 0.0  & 100.0& 66.7 & 53.6 & 57.1 & 50.0 & 86.7 & 58.9 & 92.9 & 25.0 \\[1.5pt]
AV Regurgitation, $n=40$      & 67.5 & 38.3 & 64.0 & 33.3 & 94.7 & 31.1 & 60.3 & 25.9 & 94.7 & 53.3 & 49.1 & 44.4 & 53.8 & 0.0  & 50.0 & 0.0  & 100.0& 80.6 & 50.0 &100.0 & 0.0  & 65.5 & 44.9 & 66.7 & 23.1 & 38.1 & 37.9 & 29.6 & 46.2 \\[1.5pt]
LV Syst.\ Dysfunc., $n=28$    & 60.7 & 23.8 & 60.1 & 29.4 & 90.8 & 22.7 & 59.7 & 29.4 & 89.9 & 30.0 & 58.8 & 17.6 &100.0 & 30.0 & 58.8 & 17.6 &100.0 & 0.0  & 50.0 & 0.0  & 100.0& 78.9 & 66.8 & 88.2 & 45.5 & 62.9 & 50.5 & 64.7 & 36.4 \\[1.5pt]
RV Enlargement, $n=17$        & 58.8 & 18.2 & 60.6 & 30.0 & 91.1 & 8.3  & 52.1 & 10.0 & 94.2 & 52.6 & 46.4 & 50.0 & 42.9 & 18.2 & 55.0 & 10.0 &100.0 & 0.0  & 50.0 & 0.0  & 100.0& 72.0 & 52.1 & 90.0 & 14.3 & 44.4 & 41.4 & 40.0 & 42.9 \\[1.5pt]
AV Stenosis, $n=52$           & 21.1 & 20.0 & 57.5 & 18.2 & 96.9 & 31.6 & 62.5 & 27.3 & 97.8 & 27.6 & 51.1 & 36.4 & 65.9 & 0.0  & 50.0 & 0.0  & 100.0& 0.0  & 47.6 & 0.0  & 95.1 & 27.0 & 47.1 & 45.5 & 48.8 & 28.6 & 55.1 & 27.3 & 82.9 \\[1.5pt]
\rowcolor{yellow!12}
Macro Average                 &      & 22.4 & 59.0 & 27.9 & 90.1 & 20.0 & 56.1 & 21.4 & 90.8 & 44.4 & 48.6 & 39.7 & 57.5 & 23.0 & 49.2 & 18.4 & 80.0 & 16.1 & 49.5 & 20.0 & 79.0 & 62.0 & 52.9 & 69.5 & 36.3 & 52.1 & 48.8 & 50.9 & 46.7 \\
\midrule
\multicolumn{30}{l}{\cellcolor{yellow!8}\textit{\textbf{EchoNet-Dynamic}~\cite{ouyang2020echonetdynamic} ($n = 1{,}277$, single-view)}} \\[3pt]
LV Systolic Dysfunction       &      & 39.3 & 60.2 & 31.1 & 89.1 & 37.2 & 58.7 & 27.1 & 90.3 & 5.6  & 50.6 & 3.0  & 98.2 & 51.2 & 50.5 & 98.4 & 2.5  & 66.7 & 50.0 &100.0 & 0.0  & 36.0 & 51.8 & 35.0 & 68.6 & 7.7  & 50.0 & 4.3  & 95.6 \\
\bottomrule
\end{tabular}%
}
\end{table*}
\paragraph{Per-Section Report Generation.}
~\cref{tab:green_per_section} reports Echo-GREEN scores broken down by
report section. EchoSonar-R leads on all 13 sections, with the largest margins
over baselines on valvular sections and Pulmonary Artery. Structurally simple sections such as Pericardium and Aortic Root are well-handled by most models. The Conclusions section remains the most challenging across all models, requiring integration findings across all cardiac
structures.
\begin{table*}[t]
\centering
\caption{Echo-GREEN scores per report section on the private test set.
Higher is better. EchoSonar-R$^{\dagger}$ denotes GRPO 
training; EchoSonar-R$^{\ast}$ denotes SFT-only. Best score per section in \textbf{bold}.}
\label{tab:green_per_section}
\setlength{\tabcolsep}{5pt}
\renewcommand{\arraystretch}{0.8}
\resizebox{\textwidth}{!}{%
\begin{tabular}{l cccccc c}
\toprule
\textbf{Section} 
& \textbf{Qwen3-VL} 
& \textbf{MedGemma} 
& \textbf{Chiron-o1} 
& \textbf{Lingshu} 
& \textbf{EchoSonar-R$^{\ast}$} 
& \textbf{EchoSonar-R$^{\dagger}$} \\
\midrule
Left Ventricle   & 0.394 & 0.542 & 0.427 & 0.541 & 0.758 & \textbf{0.758} \\
Right Ventricle  & 0.181 & 0.760 & 0.527 & 0.547 & 0.953 & \textbf{0.959} \\
Left Atrium      & 0.259 & 0.331 & 0.399 & 0.551 & 0.789 & \textbf{0.795} \\
Right Atrium     & 0.255 & 0.351 & 0.444 & 0.706 & 0.969 & \textbf{0.965} \\
Aortic Valve     & 0.195 & 0.117 & 0.263 & 0.403 & 0.811 & \textbf{0.817} \\
Mitral Valve     & 0.210 & 0.308 & 0.256 & 0.220 & 0.676 & \textbf{0.672} \\
Tricuspid Valve  & 0.086 & 0.226 & 0.245 & 0.276 & 0.646 & \textbf{0.658} \\
Pulmonic Valve   & 0.078 & 0.392 & 0.329 & 0.443 & 0.796 & \textbf{0.804} \\
Pericardium      & 0.371 & 0.961 & 0.493 & 0.918 & \textbf{0.988} & 0.987 \\
Aortic Root      & 0.354 & 0.920 & 0.550 & 0.754 & 0.937 & \textbf{0.936} \\
Aortic Arch      & 0.149 & 0.500 & 0.341 & 0.469 & 0.951 & \textbf{0.944} \\
Pulmonary Artery & 0.049 & 0.172 & 0.120 & 0.202 & 0.494 & \textbf{0.527} \\
Conclusions      & 0.226 & 0.311 & 0.263 & 0.354 & 0.578 & \textbf{0.578} \\
\midrule
\rowcolor{gray!6}
\textbf{Average} & 0.216 & 0.457 & 0.359 & 0.491 & 0.796 & \textbf{0.800} \\
\bottomrule
\end{tabular}}
\end{table*}

\paragraph{Qualitative Analysis.}
~\cref{fig:qualitative_mimic}-\cref{fig:qualitative_tp} show
representative EchoSonar-R predictions with full reasoning traces
across three characteristic cases. ~\cref{fig:qualitative_mimic}
illustrates a label-shift false positive on MIMICEchoQA: the model identifies trace mitral and tricuspid regurgitation, but only tricuspid regurgitation
carries a positive label under the annotation convention.

~\cref{fig:qualitative_private} shows a correct true-negative
prediction on a fully normal private study. The model reasons
systematically across six views, assessing all detected structures per
view and arriving at consistent all-absent conclusions. 

~\cref{fig:qualitative_tp} shows a largely correct true-positive
case with one false positive. The model correctly identifies LV
Systolic Dysfunction, MV Regurgitation, and TV Regurgitation, all
supported by consistent cross-view evidence of borderline LV dilation
with mild global hypokinesis. The false positive for LV Enlargement is
attributable to the same ``borderline dilated'' characterisation.

Tab.~\ref{tab:perclass_ablation} extends the aggregate ablation results with per-class F1, balanced accuracy, 
sensitivity, and specificity across all five configurations. Removing reasoning targets 
increases sensitivity at the cost of specificity (macro Sens: 47.6\%  vs.\ 43.4\%; macro Spec: 83.4\% vs.\ 86.8\%), confirming that reasoning supervision induces a more conservative decision rule. The effect is most pronounced for MV Calcification, where removing reasoning substantially increases sensitivity (71.4\% vs.\ 44.8\%) but at the cost of specificity (88.1\% vs.\ 94.7\%). Removing video tokens 
(Remove VT) collapses performance to near-random BAcc across all classes, with F1 dropping to zero for 10 of 12 conditions. The model defaults to predicting the negative class, as reflected by near-100\% specificity across all conditions. Structure tokens (Remove ST) provide a complementary contribution: the detector improves F1 on 10 of 12 classes (mean $\Delta$F1 $= +2.9\%$ across improved classes) and BAcc on 9 of 12 (mean $\Delta$BAcc $= +2.5\%$), with performance drops concentrated in diseases with the sparsest 
detector training data (RA Enlargement, Bicuspid AV).

\begin{table*}[t]
\centering
\caption{Per-class ablation results on the private test set. 
VT = video tokens; ST = structure tokens.}
\label{tab:perclass_ablation}
\setlength{\tabcolsep}{4pt}
\resizebox{\textwidth}{!}{%
\begin{tabular}{l | cccc | cccc | cccc | cccc | cccc}
\toprule
& \multicolumn{4}{c|}{\textbf{w/ reasoning}} 
& \multicolumn{4}{c|}{\textbf{w/o reasoning}} 
& \multicolumn{4}{c|}{\textbf{Remove Both}} 
& \multicolumn{4}{c|}{\textbf{Remove VT}} 
& \multicolumn{4}{c}{\textbf{Remove ST}} \\
\cmidrule(lr){2-5}\cmidrule(lr){6-9}\cmidrule(lr){10-13}
\cmidrule(lr){14-17}\cmidrule(lr){18-21}
\textbf{Abnormality} & F1 & BAcc & Sens & Spec & F1 & BAcc & Sens & Spec & F1 & BAcc & Sens & Spec & F1 & BAcc & Sens & Spec & F1 & BAcc & Sens & Spec \\
\midrule
TV Regurgitation        & 66.6 & 59.3 & 72.4 & 46.3 & 65.6 & 55.1 & 75.3 & 34.9 & 58.8 & 49.5 & 64.4 & 34.5 & 55.2 & 53.2 & 53.0 & 53.5 & 65.3 & 54.7 & 75.0 & 34.4 \\[1.5pt]
MV Regurgitation        & 69.1 & 62.6 & 75.1 & 50.0 & 70.4 & 62.8 & 78.6 & 47.0 & 57.2 & 48.2 & 62.4 & 34.1 & 56.9 & 49.5 & 60.4 & 38.6 & 65.9 & 57.9 & 72.8 & 42.9 \\[1.5pt]
Aortic Stenosis         & 69.8 & 80.9 & 63.8 & 98.0 & 59.4 & 77.3 & 58.6 & 95.9 & 0.0  & 50.0 & 0.0  & 100.0 & 0.0  & 50.0 & 0.0  & 100.0 & 66.0 & 78.6 & 59.5 & 97.8 \\[1.5pt]
MV Calcification        & 44.5 & 69.7 & 44.8 & 94.7 & 48.1 & 79.8 & 71.4 & 88.1 & 0.0  & 50.0 & 0.0  & 100.0 & 0.0  & 50.0 & 0.0  & 100.0 & 44.1 & 69.6 & 44.8 & 94.5 \\[1.5pt]
RA Enlargement          & 18.4 & 55.9 & 13.6 & 98.3 & 12.3 & 53.5 & 8.5  & 98.5 & 0.0  & 50.0 & 0.0  & 100.0 & 0.0  & 50.0 & 0.0  & 100.0 & 15.8 & 54.6 & 10.2 & 99.0 \\[1.5pt]
RV Enlargement          & 3.6  & 50.6 & 3.2  & 98.0 & 11.4 & 53.1 & 6.5  & 99.8 & 0.0  & 50.0 & 0.0  & 100.0 & 0.0  & 50.0 & 0.0  & 100.0 & 15.4 & 55.7 & 12.9 & 98.6 \\[1.5pt]
LA Enlargement          & 62.7 & 74.6 & 65.5 & 83.8 & 56.6 & 70.1 & 51.2 & 88.9 & 0.0  & 50.0 & 0.0  & 100.0 & 0.0  & 50.0 & 0.0  & 100.0 & 65.9 & 76.7 & 67.0 & 86.4 \\[1.5pt]
LV Systolic Dysfunction & 55.9 & 71.9 & 49.0 & 94.9 & 54.4 & 70.5 & 44.9 & 96.1 & 0.0  & 49.7 & 0.0  & 99.4 & 0.0  & 49.9 & 0.0  & 99.7 & 53.7 & 70.9 & 47.5 & 94.3 \\[1.5pt]
LV Enlargement          & 53.9 & 71.5 & 44.3 & 98.8 & 43.1 & 65.3 & 31.4 & 99.1 & 0.0  & 50.0 & 0.0  & 100.0 & 0.0  & 50.0 & 0.0  & 100.0 & 53.3 & 72.1 & 45.7 & 98.4 \\[1.5pt]
Bicuspid AV             & 19.2 & 57.4 & 16.1 & 98.6 & 31.6 & 68.0 & 38.7 & 97.2 & 0.0  & 50.0 & 0.0  & 100.0 & 0.0  & 50.0 & 0.0  & 100.0 & 18.2 & 56.1 & 12.9 & 99.2 \\[1.5pt]
AV Regurgitation        & 41.3 & 65.1 & 39.9 & 90.3 & 41.0 & 67.5 & 58.0 & 77.1 & 0.0  & 50.0 & 0.0  & 100.0 & 0.0  & 50.0 & 0.0  & 100.0 & 39.1 & 63.6 & 36.2 & 91.0 \\[1.5pt]
Healthy                 & 36.2 & 61.3 & 33.0 & 89.5 & 38.5 & 63.3 & 48.1 & 78.4 & 0.0  & 50.0 & 0.0  & 100.0 & 0.0  & 50.0 & 0.0  & 100.0 & 24.1 & 55.7 & 17.9 & 93.5 \\[1.5pt]
\rowcolor{yellow!12}
\textbf{Macro Average}  & 45.1 & 65.1 & 43.4 & 86.8 & 44.4 & 65.5 & 47.6 & 83.4 & 10.2 & 50.2 & 10.3 & 90.2 & 10.5 & 49.8 & 11.5 & 88.0 & 43.9 & 63.9 & 41.9 & 85.8 \\
\bottomrule
\end{tabular}%
}
\end{table*}

\begin{figure*}[t]
\centering
\begin{tcolorbox}[
    colback=gray!3, colframe=gray!40,
    colbacktitle=gray!40, coltitle=black,
    boxrule=0.4pt, arc=2pt,
    left=6pt, right=6pt, top=4pt, bottom=4pt,
    fontupper=\scriptsize,
    title={\small\rmfamily MIMICEchoQA},
]
\textbf{Available view:} A2C \quad \textbf{Detected:} LA, LV, RA, RV\\[3pt]
\textbf{Reasoning trace:}\\[2pt]
View: A2C | Detected Structures: LA, LV, RA, RV\\[2pt]
Clinical Findings:\\
\hspace*{1em}Left Ventricle (LV): Normal dimensions and systolic function.
Normal wall thickness. Diastolic filling pattern is normal.\\
\hspace*{1em}Right Ventricle (RV): Normal dimensions and systolic function.\\
\hspace*{1em}Left Atrium (LA): Normal dimensions. No suggestion of mass or
thrombus.\\
\hspace*{1em}Right Atrium (RA): Normal dimensions. No suggestion of mass or
thrombus.\\
\hspace*{1em}Aortic Valve (AV): Normal trileaflet morphology. Normal Doppler
flow characteristics.\\
\hspace*{1em}Mitral Valve (MV): Normal appearance without prolapse. Trace
mitral regurgitation.\\
\hspace*{1em}Tricuspid Valve (TV): Normal appearance. Trace tricuspid
regurgitation.\\
\hspace*{1em}Pulmonic Valve (PV): Normal appearance. Normal Doppler flow
characteristics.\\
\hspace*{1em}Pericardium: Appears normal. No significant effusion is seen.\\
\hspace*{1em}Aortic Root: Normal dimensions for BSA.\\
\hspace*{1em}Aortic Arch: Normal dimensions. Normal Doppler flow profile.\\
\hspace*{1em}Pulmonary Artery: Pulmonary artery systolic pressure could not
be estimated due to insufficient tricuspid regurgitation.\\[3pt]
Interpretation: The A2C view provides a comprehensive assessment of the
left and right ventricles, atria, and the aortic and tricuspid valves.
The normal dimensions and systolic function of the LV and RV, along with
normal wall thickness and diastolic filling patterns, suggest no
significant structural or functional abnormalities. The normal appearance
of the aortic and pulmonary valves indicates no significant valvular
disease. The presence of trace mitral and tricuspid regurgitation is
noted, which is often seen in the absence of significant structural
abnormalities and may be due to minor valve leaflet incompetence or minor
annular dilation. The normal dimensions and absence of mass or thrombus in
the atria suggest no significant atrial pathology. The inability to
estimate pulmonary artery systolic pressure due to insufficient tricuspid
regurgitation suggests that the tricuspid regurgitation is minimal and
does not significantly affect pulmonary artery pressure.\\[3pt]
Conclusions: Normal LV and RV dimensions and systolic function. Trace
mitral and tricuspid regurgitation noted, likely due to minor valve leaflet
incompetence. Normal atrial dimensions. Normal aortic root and ascending
aorta. No pericardial effusion.\\[4pt]
\rmfamily\small
\textbf{Predicted:} TV Regurgitation, MV Regurgitation\\
\textbf{Ground truth:} TV Regurgitation\\
\end{tcolorbox}
\caption{Qualitative example from MIMICEchoQA.}
\label{fig:qualitative_mimic}
\end{figure*}

\begin{figure*}[t]
\centering
\begin{tcolorbox}[
    colback=gray!3, colframe=gray!40,
    colbacktitle=gray!40, coltitle=black,
    boxrule=0.4pt, arc=2pt,
    left=6pt, right=6pt, top=4pt, bottom=4pt,
    fontupper=\scriptsize,
    title={\small\rmfamily Private set},
]
\textbf{Available views:} PLAX Standard, PSAX Zoomed Out, A4C, A4C Zoomed
LV, A2C, A3C\\[3pt]
\textbf{Reasoning trace:}\\[2pt]
View: PLAX Standard | Detected: LA, LV, LVOT\\
\hspace*{1em}LV: No significant abnormalities in internal dimensions,
septal and posterior wall thickness, global and regional systolic
function, LVOT diameter, or EPSS.\\
\hspace*{1em}LA: No significant enlargement or structural abnormalities.\\
\hspace*{1em}AV: No significant calcification, thickening, or leaflet
excursion issues.\\
\hspace*{1em}MV: No significant calcification, subvalvular apparatus
issues, or annular diameter abnormalities.\\
\hspace*{1em}Pericardium: No effusion detected.\\[2pt]
View: PSAX Zoomed Out | Detected: LA, LV, LVOT, RA, RV\\
\hspace*{1em}LV: No significant abnormalities in short-axis wall motion,
fractional shortening, regional wall motion abnormalities, or septal
flattening.\\
\hspace*{1em}AV: No significant abnormalities in valve orifice area,
number of cusps, or commissural fusion.\\
\hspace*{1em}TV: No significant tricuspid regurgitation noted.\\[2pt]
View: A4C | Detected: LA, LV, MV, RA, RV, TV\\
\hspace*{1em}LV: No significant abnormalities in endocardial tracing for
volumes, LVEF calculation, global longitudinal strain, or apical wall
motion.\\
\hspace*{1em}MV: No significant abnormalities in leaflet morphology or
mitral inflow assessment.\\
\hspace*{1em}TV: No significant tricuspid regurgitation noted.\\[2pt]
View: A4C Zoomed LV | Detected: LV\\
\hspace*{1em}LV: No significant abnormalities in LVEF calculation, global
longitudinal strain, or apical wall motion.\\[2pt]
View: A2C | Detected: LA, LV\\
\hspace*{1em}LV: No significant abnormalities in anterior and inferior
wall motion, endocardial tracing for biplane volumes, or apical
contractility.\\[2pt]
View: A3C | Detected: LV\\
\hspace*{1em}LV: No significant abnormalities in anteroseptal and
inferolateral wall motion, longitudinal contractility, or LVOT morphology.\\
\hspace*{1em}AV: No significant abnormalities in leaflet separation and
Doppler alignment.\\[3pt]
Conclusions: No significant abnormalities in any cardiac structures or
functions assessed. LV shows normal dimensions, wall thickness, and
systolic function. LA is of normal size. AV and MV show no significant
pathology. RV and RA are of normal size and function with no significant
tricuspid regurgitation. No pericardial effusion. Overall, the
echocardiogram is normal.\\[4pt]
\rmfamily\small
\textbf{Predicted:} Healthy
\textbf{Ground truth:} Healthy \\

\end{tcolorbox}
\caption{Qualitative example from the private test set.}
\label{fig:qualitative_private}
\end{figure*}

\begin{figure*}[t]
\centering
\begin{tcolorbox}[
    colback=gray!3, colframe=gray!40,
    colbacktitle=gray!40, coltitle=black,
    boxrule=0.4pt, arc=2pt,
    left=6pt, right=6pt, top=4pt, bottom=4pt,
    fontupper=\scriptsize,
    title={\small\rmfamily Private set},
]
\textbf{Available views:} PLAX Standard, PSAX Zoomed Out, A2C, A3C
\quad \textbf{Question:} What abnormalities are present?\\[3pt]
\textbf{Reasoning trace:}\\[2pt]
View: PLAX Standard | Detected: LV, LA, AV, AR, MV, RV, Pericardium\\
\hspace*{1em}LV: Borderline dilated (taking patient size into account)
with mild global hypokinesis. Possible anterior septal hypokinesis.\\
\hspace*{1em}RV: Normal dimensions and systolic function.\\
\hspace*{1em}LA: Normal dimensions. No suggestion of mass or thrombus.\\
\hspace*{1em}RA: Normal dimensions. No suggestion of mass or thrombus.\\
\hspace*{1em}AV: Normal trileaflet morphology. Normal Doppler flow
characteristics.\\
\hspace*{1em}MV: Normal appearance without prolapse. Mild mitral
regurgitation.\\
\hspace*{1em}TV: Normal appearance. Mild-to-moderate tricuspid
regurgitation.\\
\hspace*{1em}Pericardium: Appears normal. No significant effusion.\\
\hspace*{1em}Aortic Root: Normal dimensions for BSA.\\
\hspace*{1em}Pulmonary Artery: Systolic pressure estimated to be normal.\\
\hspace*{1em}Implications: The borderline dilated LV with mild global
hypokinesis and possible anterior septal hypokinesis suggests early
systolic dysfunction, possibly related to coronary artery disease or
hypertension. The mild MV and TV regurgitation indicate mild valvular
insufficiency, possibly secondary to LV dysfunction.\\[2pt]
View: PSAX Zoomed Out | Detected: LV, LA, AV, MV, RV, TV, RA, Pericardium\\
\hspace*{1em}LV: Borderline dilated with mild global hypokinesis.
Possible anterior septal hypokinesis. Findings consistent with PLAX
view, reinforcing the presence of mild global hypokinesis.\\
\hspace*{1em}AV: Normal trileaflet morphology. Normal Doppler flow
characteristics.\\
\hspace*{1em}MV: Normal appearance without prolapse. Mild mitral
regurgitation.\\
\hspace*{1em}TV: Normal appearance. Mild-to-moderate tricuspid
regurgitation.\\
\hspace*{1em}Pericardium: No significant effusion.\\[2pt]
View: A2C | Detected: LV, LA, MV\\
\hspace*{1em}LV: Borderline dilated with mild global hypokinesis.
Possible anterior septal hypokinesis. Findings consistent with the
PLAX and PSAX views, reinforcing mild global hypokinesis.\\
\hspace*{1em}LA: Normal dimensions. No suggestion of mass or thrombus.\\
\hspace*{1em}MV: Normal appearance without prolapse. Mild mitral
regurgitation.\\[2pt]
View: A3C | Detected: LV, AV, LA, AA, MV, LVOT\\
\hspace*{1em}LV: Borderline dilated with mild global hypokinesis.
Possible anterior septal hypokinesis. Anteroseptal and inferolateral
wall motion consistent with prior views.\\
\hspace*{1em}AV: No significant abnormalities in leaflet separation or
Doppler alignment.\\
\hspace*{1em}MV: Normal coaptation point. No mitral-aortic curtain
calcification.\\[3pt]
\rmfamily\small
\textbf{Predicted:} TV Regurgitation, MV Regurgitation, LV Systolic Dysfunction, LV Enlargement\\
\textbf{Ground truth:} TV Regurgitation, MV Regurgitation, LV Systolic
Dysfunction
\end{tcolorbox}
\caption{Qualitative example from the private test set.}
\label{fig:qualitative_tp}
\end{figure*}
\end{document}